%% file: arxiv.tex
\documentclass{article}
\usepackage{ijcai26}

\usepackage{times}
\usepackage{soul}
\usepackage{url}
\usepackage[hidelinks]{hyperref}
\usepackage[utf8]{inputenc}
\usepackage[small]{caption}
\usepackage{subcaption}
\usepackage{graphicx}
\usepackage{amsfonts}    
\usepackage{amsmath}
\usepackage{amsthm}
\usepackage{booktabs}
\usepackage{algorithm}
\usepackage{algorithmic}
\usepackage[switch]{lineno}
\usepackage{makecell}

\allowdisplaybreaks
\makeatletter
\newtheorem*{rep@theorem}{\rep@title}
\newcommand{\newreptheorem}[2]{%
\newenvironment{rep#1}[1]{%
 \def\rep@title{#2 \ref{##1}}%
 \begin{rep@theorem}}%
 {\end{rep@theorem}}}
\makeatother

\usepackage{stackengine}
\def\delequal{\mathrel{\ensurestackMath{\stackon[1pt]{=}{\scriptstyle\Delta}}}}

\newtheorem{theorem}{Theorem}
\newreptheorem{theorem}{Theorem}
\newtheorem{definition}[theorem]{Definition}

\title{Identification of Probabilities of Causation: from Recursive to Closed-Form Bounds}

\author{
Xin Shu
\and
Shuai Wang\and
Ang Li$^*$\\
\affiliations
Department of Computer Science, Florida State University\\
\emails
xs24a@fsu.edu, sw23v@fsu.edu,
angli@cs.fsu.edu
}

\begin{document}

\maketitle

\begin{abstract}
    Probabilities of causation (PoCs) are fundamental quantities for counterfactual analysis and personalized decision making. However, existing analytical results are largely confined to binary settings. This paper extends PoCs to multi-valued treatments and outcomes by deriving closed form bounds for a representative family of discrete PoCs within Structural Causal Models, using standard experimental and observational distributions. We introduce the notion of equivalence classes of PoCs, which reduces arbitrary discrete PoCs to this family, and establish a replaceability principle that transfers bounds across value permutations. For the resulting bounds, we prove soundness in all dimensions and empirically verify tightness in low dimensional cases via Balke's linear programming method; we further conjecture that this tightness extends to all dimensions. Simulations indicate that our closed form bounds consistently tighten recent recursive bounds while remaining simpler to compute. Finally, we illustrate the practical relevance of our results through toy examples.
\end{abstract}

\section{Introduction}

In fields such as healthcare, finance, and energy, probabilities of causation can enable effective, accurate, and explainable decision-making in complex scenarios. This causality-based approach addresses key challenges, including selection bias, counterfactual reasoning, individualized decision-making, and proper credit assignment—ensuring that outcomes are attributed to the treatments or factors that actually caused them. For example, Li and Pearl \shortcite{li2019unit} proposed a unit selection framework that uses a linear combination of probabilities of causation to identify individuals who are most likely to respond as desired, while also evaluating the associated rewards and costs. Similarly, Mueller and Pearl \shortcite{mueller:pea-r513} demonstrated how probabilities of causation can guide personalized medical decisions in alignment with the ethical principle of ``do no harm.''

The development of probabilities of causation began in the early 2000s, when Pearl \shortcite{pearl1999probabilities} first introduced three key binary measures: the probability of necessity (PN), the probability of sufficiency (PS), and the probability of necessity and sufficiency (PNS) in the Structural Causal Models (SCMs) \cite{galles1998axiomatic,halpern2000axiomatizing}. Tian and Pearl \shortcite{tian2000probabilities} then derived tight bounds for these quantities using Balke's linear programming method \cite{balke1995probabilistic}. Decades later, Li and Pearl \shortcite{li2019unit,li2022unit} formally proved the validity of these bounds. Building on this foundation, Muller et al. \shortcite{mueller2021causes}, along with Dawid et al. \shortcite{dawid2017probability}, further narrowed the bounds by incorporating covariate information and leveraging underlying causal structures.

The aforementioned studies are all restricted to binary treatments and outcomes, which poses significant limitations for real-world applications. In contrast, extending probabilities of causation to the non-binary case enables a more detailed and informative assessment. For example, in a binary setting, physicians can only determine whether a medication is effective or not, leading to a simple yes-or-no prescription decision. However, with multi-valued treatments $x_i$ (e.g., different dosage levels, where $1 \le i \le n$ and 
$n \in \mathbb{N}$), doctors can evaluate which dosage level yields the optimal therapeutic effect for a particular patient. This finer-grained understanding supports more personalized and effective treatment decisions.

Efforts to address non-binary cases have continued in recent work. Zhang et al.~\shortcite{zhang2022partial}, as well as Li and Pearl \shortcite{li2022bounds}, proposed numerical methods for computing non-binary probabilities of causation using nonlinear programming techniques. 

More recently, Li and Pearl \shortcite{li2024probabilities} introduced recursive-form but un-tight theoretical bounds for non-binary probabilities. Relatedly, de Aguas et al. \shortcite{de2025probability} has studied probabilities of tiered benefit and harm for ordinal outcomes, focusing on cross-threshold causal effects and semiparametric inference. However, numerical methods are constrained by computational complexity, and recursive forms often lack sharpness, particularly in high-dimensional settings, indicating that this line of research still requires substantial advancement.

In this paper, we present non-recursive theoretical bounds for arbitrary non-binary probabilities of causation without imposing any structural restrictions.
We further verified both the soundness and completeness of these bounds empirically using Balke’s linear programming framework \cite{balke1995probabilistic} in low-dimensional settings (specifically when the treatment and outcome take three or four possible values).
For the general n-dimensional case, we provide a formal proof of soundness showing that the proposed formulas indeed define valid lower and upper bounds, while a complete proof establishing their tightness remains open.
Our approach first identifies a small set of powerful non-binary probabilities of causation that are sufficient to represent any others, and then derives bounds for this representative set.

\subsection{Key Contributions}
\begin{itemize}
    \item We derive closed-form bounds for multi-valued PoCs in general SCMs using only experimental and observational data.
    \item We introduce equivalence classes and a replaceability principle to reduce arbitrary PoC queries and transfer bounds across value permutations.
    \item We prove soundness for all dimensions, verify tightness in low dimensions, and show consistent tightening over the Li--Pearl method with improved computational simplicity.
\end{itemize}

\section{Preliminaries}

We begin by reviewing the basic definitions of probabilities of causation and the associated notation in Structural Causal Models (SCMs) \cite{galles1998axiomatic,halpern2000axiomatizing}. Readers already familiar with SCMs may choose to skip this section.

Counterfactuals are well defined within the SCM framework. A counterfactual statement such as ``Variable $Y$ would have the value $y$ had $X$ been $x$'' is denoted as $Y_x = y$. Here, the subscript $x$ refers to a hypothetical event, which is well-defined in SCMs. For brevity, we will use $y_x$, $y_{x'}$, $y'_x$, and $y'_{x'}$ throughout the paper to represent the events $Y_x = y$, $Y_{x'} = y$, $Y_x = y'$, and $Y_{x'} = y'$, respectively. Unless otherwise specified, we assume that experimental data are available in the form of causal effects (e.g., $P(y_x)$), while observational data are provided as joint probabilities (e.g., $P(x, y)$). In our notation, $X$ denotes the treatment variable and $Y$ denotes the outcome or effect.

Let $X$ and $Y$ be two binary variables in a causal model $M$, where $x$ denotes the proposition $X = \text{true}$, $x'$ denotes $X = \text{false}$, $y$ denotes $Y = \text{true}$, and $y'$ denotes $Y = \text{false}$. The three basic binary probabilities of causation—PN, PS, and PNS—are defined as follows \cite{pearl1999probabilities}:

\begin{definition}[Probabilities of Causation] \cite{pearl1999probabilities}
\begin{align*}
\text{PN}  \delequal  P(y'_{x'}|x,y),
\text{PS} \delequal P(y_x|y',x'),
\text{PNS}\delequal P(y_x,y'_{x'})
\end{align*}
\end{definition}

PNS stands for the probability that $y$ would respond to $x$ both ways, and therefore measures both the sufficiency and necessity of $x$ to produce $y$.

Tian and Pearl \shortcite{tian2000probabilities} then derived tight bounds for the PN, PS, and PNS using Balke's linear programming approach \cite{balke1995probabilistic}. These bounds were later given a formal theoretical proof by Li and Pearl~\shortcite{li2019unit,li2022unit}.

The tight bounds for binary PNS, PN, and PS, are given as follows:
\begin{align*}
\max \left \{
\begin{array}{cc}
0, \\
P(y_x) - P(y_{x'}), \\
P(y) - P(y_{x'}), \\
P(y_x) - P(y)
\end{array}
\right \}
\le \text{PNS}
\end{align*}

\begin{align*}
\text{PNS}
\le \min \left \{
\begin{array}{cc}
 P(y_x), \\
 P(y'_{x'}), \\
P(x,y) + P(x',y'), \\
P(y_x) - P(y_{x'}) + P(x, y') + P(x', y)
\end{array} 
\right \}
\end{align*}

\begin{align*}
\max \left \{
\begin{array}{cc}
0, \\
\displaystyle
\frac{P(y)-P(y_{x'})}{P(x,y)}
\end{array} 
\right \} \le
\text{PN} 
\end{align*}

\begin{align*}
\text{PN} \le
\min \left \{
\begin{array}{cc}
1, \\
\displaystyle
\frac{P(y'_{x'})-P(x',y')}{P(x,y)}
\end{array}
\right \}
\end{align*}

\begin{align*}
\max \left \{
\begin{array}{cc}
0, \\
\displaystyle
\frac{P(y')-P(y'_{x})}{P(x',y')}
\end{array} 
\right \} \le
\text{PS} 
\end{align*}

\begin{align*}
\text{PS} \le
\min \left \{
\begin{array}{cc}
1, \\
\displaystyle
\frac{P(y_{x})-P(x,y)}{P(x',y')}
\end{array}
\right \}
\end{align*}

Regarding non-binary probabilities of causation, Li and Pearl \shortcite{li2024probabilities} recently provided eight theorems establishing theoretical (though not tight) recursive-form bounds for cases where both the treatment and the effect are non-binary. Specifically, when the treatment variable $X$ takes $m$ values and the outcome variable $Y$ takes $n$ values, the following set of non-binary probabilities of causation are defined and bounded. (We omit the detailed bounds here; interested readers may refer to \cite{li2024probabilities} for full derivations.)
\begin{align*}
&\textbf{Probability of preservation(i, j): } \\& P({y_i}_{x_j}, y_i),  \quad s.t., 1 \le i \le n, 1 \le j \le m 
\\
&\textbf{Probability of replacement(i, j, k): } \\& P({y_i}_{x_j}, y_k), \quad s.t., 1 \le i,k \le n, 1 \le j \le m, i\ne k\\
&\textbf{Probability of substitute(i, j, k): } \\& P({y_i}_{x_j}, x_k), \quad s.t., 1 \le i \le n, 1 \le j,k \le m, j\ne k\\
&\textbf{Probability of necessity(i, j, k, p): } \\& P({y_i}_{x_j}, y_k, x_p), \quad s.t., 1 \le i,k \le n, 1 \le j,p \le m, j\ne p\\
&\textbf{Probability of necessity and sufficiency(k):}\\&P({y_{i_1}}_{x_{j_1}},...,{y_{i_k}}_{x_{j_k}}), \quad s.t., 1 \le i_1,...,i_k \le n, \\&1 \le j_1,...,j_k \le m, j_1\ne ... \ne j_k\\
&\textbf{Probability of substitute(k,p):}\\&P({y_{i_1}}_{x_{j_1}},...,{y_{i_k}}_{x_{j_k}},x_p), \quad s.t., 1 \le i_1,...,i_k \le n, \\&1 \le j_1,...,j_k, p \le m, j_1\ne ... \ne j_k \ne p\\
&\textbf{Probability of replacement(k,q):}\\&P({y_{i_1}}_{x_{j_1}},...,{y_{i_k}}_{x_{j_k}},y_q), \quad s.t., 1 \le i_1,...,i_k,q \le n, \\&1 \le j_1,...,j_k \le m, j_1\ne ... \ne j_k\\
&\textbf{Probability of necessity(k, p, q):}\\&P({y_{i_1}}_{x_{j_1}},...,{y_{i_k}}_{x_{j_k}},x_p,y_q),\quad s.t., 1 \le i_1,...,i_k,q \le n, 
\\& 1 \le j_1,...,j_k, p \le m, j_1\ne ... \ne j_k \ne p.
\end{align*}
In this paper, we will provide tighter bounds for a subset of the modified probabilities defined above and demonstrate that these selected probabilities are sufficient to represent all discrete probabilities of causation within SCMs.

\section{Main Results}
In this section, we first simplify the definitions of the Probability of Necessity and Sufficiency(k), Probability of Substitute(k,p), Probability of Replacement(k,q), and Probability of Necessity(k,p,q) defined by Li and Pearl \shortcite{li2024probabilities} into their most concise forms, while retaining their ability to represent all discrete probabilities of causation in Structural Causal Models (SCMs).

\subsection{Probability of necessity and sufficiency($k$)}
The Probability of Necessity and Sufficiency($k$) is simplified to the form $P({y_{1}}_{x_{1}}, \ldots, {y_{k}}_{x_{k}})$, where $|X| = |Y| = n$, and the values of $X$ and $Y$ are arranged in increasing order within the probability expression. This simplification yields the most straightforward closed-form representation. We will later demonstrate that this simplification preserves generality. The closed-form bounds for the modified Probability of Necessity and Sufficiency($k$) are presented in the following theorem. All theorem proofs are provided in the appendix.

\begin{theorem}[Probability of necessity and sufficiency($k$) (PNS($k$))]\label{nnk}
Suppose variable $X$ has $n$ values $x_1,...,x_n$ and $Y$ has $n$ values $y_1,...,y_n$, $k \le n$, then the probability of necessity and sufficiency($k$) $P({y_{1}}_{x_{1}},...,{y_{k}}_{x_{k}})$ is bounded as following:
\begin{align*}
\max \left \{
\begin{array}{cc}
0, \\
\displaystyle \sum_{j = 1}^{k}P({y_{j}}_{x_{j}}) - k + 1, \\
\displaystyle \sum_{\substack{1 \le j \le k \\ j \ne i}}\left[P({y_{j}}_{x_{j}})+P({x_{j}})-P({x_{j}, y_{j}})\right]+ \\
+ P({x_{i}, y_{i}}) - k + 1, \quad i \in \{1, ..., k\}
\end{array}
\right \}\nonumber\\
\le P({y_{1}}_{x_{1}},...,{y_{k}}_{x_{k}})
\label{nnklb}
\end{align*}

\begin{align*}
\min \left \{
\begin{array}{cc}
\displaystyle \sum_{j = 1}^{k}P({x_{j}, y_{j}}) + \sum_{j = k+1}^{n}P({x_{j}}), \\
P({y_{j}}_{x_{j}}), \qquad j \in \{1,...,k\},\\
\displaystyle \frac{1}{m}\left[\sum_{j = 0}^{m}P({y_{t_j}}_{x_{t_j}}) - P({x_{t_j}, y_{t_j}})\right], \\\qquad\qquad m\in\{1,...,k-1\}, t_j\in\{1,...,k\}
\end{array} 
\right \}\nonumber
\\ \ge P({y_{1}}_{x_{1}},...,{y_{k}}_{x_{k}})
\end{align*}
\end{theorem}

Note that PNS$(k)$ is a nonbinary, higher-order generalization of the original PNS. For example, PNS$(2)$—which reduces to the binary PNS—represents the probability that $y_1$ would respond to $x_1$ and $y_2$ would respond to $x_2$, formally written as $P({y_1}_{x_1}, {y_2}_{x_2})$. Bounds for similar probabilities, such as $P({y_2}_{x_1}, {y_1}_{x_2})$, can also be derived by leveraging the replaceability property of the probabilities of causation, as described in Theorem~\ref{nnk_replacement}.

The proof of soundness, specifying when the bounds are attained with equality, is provided in the appendix. We will also present toy examples to illustrate how our results enhance decision-making capabilities. While some may argue that the previous definition of PNS$(k)$ is more general, we will later demonstrate their equivalence in expressive power in Theorem~\ref{nnk_mnk}.

\subsection{Probability of substitute($k,p$)}
Similar to the Probability of Necessity and Sufficiency($k$), the Probability of Substitute($k, p$) is simplified as stated in the following theorem:
\begin{theorem}[Probability of substitute($k,p$) (PSub($k,p$))]\label{nnk+x_p}
Suppose variable $X$ has $n$ values $x_1,...,x_n$ and $Y$ has $n$ values $y_1,...,y_n$, $k \le n$, then the probability $P({y_{1}}_{x_{1}},...,{y_{k}}_{x_{k}}, x_p)$, s.t., $p \ne j$ for $1\le j \le k$ is bounded as following:
\begin{align*}
\max \left \{
\begin{array}{cc}
0, \\
\displaystyle \sum_{j=1}^{k}\left[P({y_{j}}_{x_{j}})+P({x_{j}})-P({x_{j}, y_{j}})\right] \\+ P({x_{p}}) - k
\end{array}
\right \}\nonumber
\\ \le P({y_{1}}_{x_{1}},...,{y_{k}}_{x_{k}}, x_p)
\end{align*}

\begin{align*}
\min \left \{
\begin{array}{cc}
P({x_{p}}),\\
P({y_{j}}_{x_{j}}) - P({x_{j}, y_{j}}), & j \in \{1,...,k\}\\
\end{array} 
\right \}\nonumber
\\ \ge P({y_{1}}_{x_{1}},...,{y_{k}}_{x_{k}}, x_p)\\
\end{align*}
\end{theorem}

\subsection{Probability of replacement($k,q$) }
Then, the Probability of replacement($k,q$) is simplified as stated in the following theorem:

\begin{theorem}[Probability of replacement($k,q$) (PRep($k,q$))]\label{nnk+y_q}
Suppose variable $X$ has $n$ values $x_1,...,x_n$ and $Y$ has $n$ values $y_1,...,y_n$, $k \le n$, then the probability $P({y_{1}}_{x_{1}},...,{y_{k}}_{x_{k}}, y_q)$ is bounded as following:
\begin{align*}
\max \left \{
\begin{array}{cc}
0, \\
\displaystyle \sum_{j=1}^{k}\left[P({y_{j}}_{x_{j}})+P({x_{j}})-P({x_{j}, y_{j}})\right] +\\
+\displaystyle \sum_{\substack{k+1 \le j \le n \\ j \ne q}}{P({x_{j}, y_{q}})} + P({x_{q}, y_{q}}) - k,\\
\displaystyle \sum_{\substack{1 \le j \le k \\ j \ne q}}\left[P({y_{j}}_{x_{j}})+P({x_{j}})-P({x_{j}, y_{j}})\right] +\\
+ P({x_{q}, y_{q}}) - (k-1), \quad \text{if } q \in \{1,...,k\} 
\end{array}
\right \}\nonumber
\\ \le P({y_{1}}_{x_{1}},...,{y_{k}}_{x_{k}}, y_q)
\end{align*}

\begin{align*}
\min \left \{
\begin{array}{cc}
P({y_{q}}_{x_{q}}), \qquad \text{if } q \in \{1,...,k\}, \\
\displaystyle P({x_{q}, y_{q}}) + \sum_{\substack{k+1 \le j \le n \\ j \ne q}}P({x_{j}, y_{q}}),\\
P({y_{j}}_{x_{j}}) - P({x_{j}, y_{j}}), \qquad j \in \{1,...,k\}, j \ne q\\
\end{array} 
\right \}\nonumber
\\ \ge P({y_{1}}_{x_{1}},...,{y_{k}}_{x_{k}}, y_q)\\
\end{align*}
\end{theorem}

\subsection{Probability of necessity($k,p,q$)}
Finally, the Probability of necessity($k,p,q$) is simplified as stated in the following theorem:
\begin{theorem}[Probability of necessity($k,p,q$) (PN($k,p,q$))]\label{nnk+x_p+y_q}
Suppose variable $X$ has $n$ values $x_1,...,x_n$ and $Y$ has $n$ values $y_1,...,y_n$, $k \le n$, then the probability $P({y_{1}}_{x_{1}},...,{y_{k}}_{x_{k}}, x_p, y_q)$, s.t., $p \ne j$ for $1\le j \le k$ is bounded as following:
\begin{align*}
\max \left \{
\begin{array}{cc}
0, \\
\displaystyle \sum_{j=1}^{k}\left[P({y_{j}}_{x_{j}})+P({x_{j}})-P({x_{j}, y_{j}})\right]+ \\ + P({x_{p}, y_{q}}) - k\\
\end{array}
\right \}\nonumber
\\ \le P({y_{1}}_{x_{1}},...,{y_{k}}_{x_{k}}, x_p, y_q)
\end{align*}

\begin{align*}
\min \left \{
\begin{array}{cc}
P({x_{p}},{y_{q}}), \\
P({y_{j}}_{x_{j}}) - P({x_{j}, y_{j}}), & j \in \{1,...,k\} \\
\end{array} 
\right \}\nonumber
\\ \ge P({y_{1}}_{x_{1}},...,{y_{k}}_{x_{k}}, x_p, y_q)
\end{align*}
\end{theorem}

\subsection{Equivalence Class in Probability of Causation}
We now proceed to demonstrate the generality of the above modified probabilities of causation through the next two theorems.

\begin{theorem}[Equivalence classes in probabilities of causation]\label{nnk_mnk}
Suppose variable $X$ has $n$ values $x_1,...,x_n$, $Y$ has $m$ values $y_1,...,y_m$:

\begin{itemize}
\item \textbf{Case 1:} For $k \le m<n$. Let $Y'$ have $n$ values $y'_1,...,y'_n$.  \\
Then the bounds of the probability, $P({y_{1}}_{x_{1}},...,{y_{k}}_{x_{k}})$, is exactly the same as the bounds of the probability, $P({y'_{1}}_{x_{1}},...,{y'_{k}}_{x_{k}})$, where
\begin{align*}
&P({y'_{l}}_{x_j}) = 0, P(y'_l)=0, \\ &\text{ for }m+1 \le l \le n, 1 \le j \le n,
\end{align*}
and,
\begin{align*}
&P({y'_{l}}_{x_j}) = P({y_{l}}_{x_j}), P(x_j, y'_l)=P(x_j, y_l),\\ &\text{ for }1 \le l \le m, 1 \le j \le n.
\end{align*}
\item \textbf{Case 2:} For $k \le n<m$. Let $X'$ have $m$ values $x'_1,...,x'_m$.  \\
Then the bounds of the probability, $P({y_{1}}_{x_{1}},...,{y_{k}}_{x_{k}})$, is exactly the same as the bounds of the probability, $P({y_{1}}_{x'_{1}},...,{y_{k}}_{x'_{k}})$, where
\begin{align*}
&P({y_{j}}_{x'_l}) = 0, P({y_{m}}_{x'_l}) = 1, P(x'_l) = 0, \\&\text{ for }n+1 \le l \le m, 1 \le j \le m-1,
\end{align*}
and,
\begin{align*}
&P({y_{j}}_{x'_l}) = P({y_{j}}_{x_l}), P(x'_l, y_j) = P(x_l, y_j) \\&\text{ for }1 \le l \le n, 1 \le j \le m.
\end{align*}
\end{itemize}
\end{theorem}

\subsection{Replaceability in Probability of Causation}

\begin{theorem}[Replaceability in probabilities of causation]\label{nnk_replacement}
Suppose variable $X$ has $n$ values $x_1,...,x_n$ and $Y$ has $n$ values $y_1,...,y_n$, then the bounds of the probability, $P({y_{1}}_{x_{1}},...,{y_{i-1}}_{x_{i-1}},{y_{\hat{i}}}_{x_{i}},{y_{i+1}}_{x_{i+1}},...,{y_{k}}_{x_{k}})$,  can be obtained by replacing ${y_i}_{x_i}$ with ${y_{\hat{i}}}_{x_i}$ for any $i$, such that $1 \le i \le n$, in the bounds of the probability, $P({y_{1}}_{x_{1}},...,{y_{k}}_{x_{k}})$.
\end{theorem}
Note that both theorems apply to the cases of PNS($k$), Psub($k, p$), PRep($k, q$), and PN($k, p, q$). For instance, to derive the bounds for $P(y_x, y_{x'}, y''_{x''} \mid x''', y''')$, we can first apply PN($3, 4, 4$) to obtain $P(y_x, y'_{x'}, y''_{x''}, x''', y''')$. If $|X| \ne |Y|$, we should first apply Theorem \ref{nnk_mnk}. Then, using the replaceability property of probabilities of causation (Theorem \ref{nnk_replacement}), we can convert $P(y_x, y'_{x'}, y''_{x''}, x''', y''')$ into $P(y_x, y_{x'}, y''_{x''}, x''', y''')$. Finally, dividing by $P(x''', y''')$ yields the desired conditional probability: $P(y_x, y_{x'}, y''_{x''} \mid x''', y''')$.

\section{Examples}

In this section, we will discuss several examples to demonstrate the effectiveness and usefulness of our results.

\subsection{Cross-Treatment Heterogeneity in Medicine}

Consider patients with hypertension who are eligible for three alternative treatment options. Standard clinical summaries report average improvement rates for each treatment, yet they fail to identify patients who would benefit from one treatment but be harmed by another. Since the same patient cannot receive all treatments simultaneously, and responses observed at different times may be confounded by disease progression, such cross-treatment responses are inherently counterfactual, motivating the use of multi-valued probabilities of causation.

Let $X$ denote the treatment assigned to a patient, where $x_1$, $x_2$, and $x_3$ correspond to Treatment~1, Treatment~2, and Treatment~3, respectively. Let $Y$ denote the clinical outcome category, with $y_1$, $y_2$, and $y_3$ indicating poor outcome, moderate outcome, and good outcome.

We focus on the probability of causation
$P({y_3}_{x_1}, {y_1}_{x_2}, {y_2}_{x_3})$,
which quantifies the fraction of patients who would achieve a good outcome under Treatment~1, a poor outcome under Treatment~2, and only a moderate outcome under Treatment~3. Such cross-treatment response patterns cannot be inferred from marginal outcome rates alone.

To obtain the required estimates, clinicians conduct a randomized clinical trial with $900$ patients, where each patient is assigned to one of the three treatments. The experimental data are summarized in Table~\ref{tb1_med}.

\begin{table}[t]
\centering
\begin{tabular}{lrrr}
\toprule
 & Treatment 1 & Treatment 2 & Treatment 3 \\
\midrule
poor      & 46  & 270 & 40  \\
moderate  & 23  & 8   & 223 \\
good      & 231 & 22  & 37  \\
\midrule
Overall   & 300 & 300 & 300 \\
\bottomrule
\end{tabular}
\caption{Experimental data collected by clinicians.}
\label{tb1_med}
\end{table}

Notably, each marginal experimental probability appearing in the target conjunction, namely
$P({y_3}_{x_1})$, $P({y_1}_{x_2})$, and $P({y_2}_{x_3})$, exceeds $0.5$.
This might suggest that the joint probability $P({y_3}_{x_1}, {y_1}_{x_2}, {y_2}_{x_3})$ is also large.
However, this inference is not justified without causal bounds that account for cross-world dependencies.

In addition, clinicians collected observational data from $900$ patients whose treatment choices were self-selected, as summarized in Table~\ref{tb2_med}.

\begin{table}[b]
\centering
\begin{tabular}{lrrr}
\toprule
 & Treatment 1 & Treatment 2 & Treatment 3 \\
\midrule
poor     & 131 & 45  & 38  \\
moderate & 68  & 22  & 483 \\
good     & 1   & 51  & 61  \\
\midrule
Overall  & 200 & 118 & 582 \\
\bottomrule
\end{tabular}
\caption{Observational data collected by clinicians.}
\label{tb2_med}
\end{table}

By plugging both the experimental and observational data into Theorem~\ref{nnk} together with the replaceability property (Theorem~\ref{nnk_replacement}), we obtain the following bounds:
\begin{eqnarray*}
0.509 \le P({y_3}_{x_1}, {y_1}_{x_2}, {y_2}_{x_3}) \le 0.588.
\end{eqnarray*}

In contrast, the recursive bounds obtained using the method of Li and Pearl~\shortcite{li2024probabilities} yield $[0.428, 0.588]$,
which leaves substantial ambiguity as to whether this response type constitutes a majority of patients.

Our tighter lower bound resolves this ambiguity by showing that more than half of patients exhibit this cross-treatment response pattern, indicating a large subgroup for which Treatment~1 is beneficial, Treatment~2 is harmful, and Treatment~3 is intermediate. Such risk-relevant heterogeneity is invisible to marginal outcome comparisons and highlights how tighter multi-valued PoC bounds can guide heterogeneity-aware treatment selection.

\subsection{Personalized Decision-Making in Educational Interventions}

An education platform provides academic support at multiple intensity levels for learners preparing for certification exams. While the platform advertises that increased instructional intensity leads to better learning outcomes on average, advisors are concerned that overly aggressive intervention may harm some learners.

Consider a learner with stable but unexceptional performance. The advisor seeks to assess whether moderate support would be beneficial, while intensive support could instead be detrimental. Such questions are central to personalized education policy, where one-size-fits-all strategies may be suboptimal.

Let $X$ denote the level of academic support, ranging from intensive ($x_1$), moderate ($x_2$), light ($x_3$) to none ($x_4$).
Let $Y$ denote the learner's grade at the end of the term, where $y_1$ indicates fail the exam (e.g., grade F), $y_2$ indicates pass the exam, and $y_3$ indicates excellent performance (e.g., grade A).

The advisor’s query is captured by the probability of causation
$P({y_1}_{x_1}, {y_2}_{x_2}, {y_2}_{x_3} \mid x_4, y_2)$,
which asks: among learners who currently pass without support,  what is the probability that they would still pass under light or moderate support, but fail under intensive intervention?

To answer this question, the advisor obtains historical data collected by the platform over several academic terms, during which learners participated in different intensity levels while following comparable curricula and assessment standards. The data include both experimental studies (where 1200 learners were assigned to different intervention levels) and observational records (where 1200 learners self-selected their level of support). These results are summarized in Tables~\ref{tb3} and~\ref{tb4}.

\begin{table}[t]
\centering
\begin{tabular}{lrrrr}
\toprule
 & Intensive & Moderate & Light & None \\
\midrule
Fail        & 195 & 11  & 80  & 100 \\
Pass        & 51  & 266 & 198 & 147 \\
Excellent   & 54  & 23  & 22  & 53  \\
\midrule
Overall     & 300 & 300 & 300 & 300 \\
\bottomrule
\end{tabular}
\caption{Experimental data summarized by the education platform.}
\label{tb3}
\end{table}

\begin{table}[b]
\centering
\begin{tabular}{lrrrr}
\toprule
 & Intensive & Moderate & Light & None \\
\midrule
Fail        & 67  & 11  & 53  & 46  \\
Pass        & 129 & 17  & 53  & 436 \\
Excellent   & 193 & 87  & 70  & 38  \\
\midrule
Overall     & 389 & 115 & 176 & 520 \\
\bottomrule
\end{tabular}
\caption{Observational data summarized by the education platform.}
\label{tb4}
\end{table}



Based on an observational study, the platform reports that higher levels of intervention are associated with improved performance on average: $16.1\%$, $7.3\%$, $5.8\%$, and $3.2\%$ of learners achieve excellent performance under intensive instruction, moderate instruction, light instruction, and no additional support, respectively, suggesting that intensive training may be beneficial. However, such averages do not address individual-level risks.

We therefore consider:
\[
P({y_1}_{x_1}, {y_2}_{x_2}, {y_2}_{x_3} \mid x_4, y_2)
= \frac{P({y_1}_{x_1}, {y_2}_{x_2}, {y_2}_{x_3}, x_4, y_2)}{P(x_4, y_2)},
\]
which quantifies the probability that increasing intervention intensity would negatively affect learners who are currently stable without support.

Applying the experimental and observational estimates to Theorem~\ref{nnk+x_p+y_q} and Theorem~\ref{nnk_replacement}, we obtain:
\begin{eqnarray*}
0.0125 \le P({y_1}_{x_1}, {y_2}_{x_2}, {y_2}_{x_3}, x_4, y_2) \le 0.363,
\end{eqnarray*}
which is strictly tighter than the bounds obtained using the method of Li and Pearl~\shortcite{li2024probabilities}: $[0,0.363]$.

Since \( P(x_4, y_2) = \frac{436}{1200} \), this yields:
\begin{eqnarray*}
0.344 \le P({y_1}_{x_1}, {y_2}_{x_2}, {y_2}_{x_3} \mid x_4, y_2) \le 1,
\end{eqnarray*}
which is substantially more informative than the vacuous bound $[0,1]$ obtained by Li and Pearl~\shortcite{li2024probabilities} method.

These results allow academic advisors to conclude that, for a non-negligible subset of learners, intensive intervention poses a significant risk of adverse outcomes, while moderate support remains safe.

\section{Simulated Results}
In this section, we empirically demonstrate that the conjectured closed-form bounds
yield tighter estimates than the recursive bounds of Li and Pearl~\shortcite{li2024probabilities}.
Here we focus on the probability of necessity and sufficiency  ($\mathrm{PNS}$) to provide a simple and visually direct illustration of how much improvement is obtained in practice.

\begin{figure}[!b]
  \centering
  \includegraphics[width=0.85\linewidth]{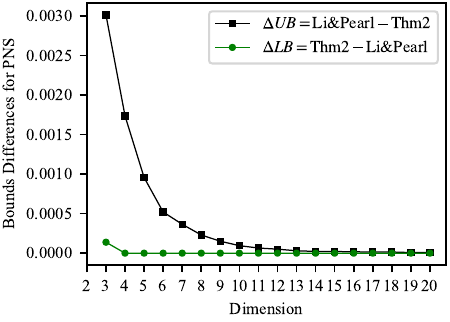}
  \caption{Bounds differences between Li--Pearl and Theorem~2.}
  \label{pic3}
\end{figure}

\paragraph{Dimension sweep ($n\in[3,20]$).}
We fix $k=3$ (i.e., $X$ and $Y$ each take three values) and vary the dimensionality
$n\in[3,20]$, focusing on the probability of necessity and sufficiency $\mathrm{PNS}(k)$.
For each $n$, we randomly generate $10{,}000$ ground-truth instances of $\mathrm{PNS}(3)$,
i.e., $P({y_1}_{x_1},{y_2}_{x_2},{y_3}_{x_3})$.
For every instance, we construct compatible observational and experimental sample distributions.
Our data-generation procedure enforces the standard feasibility constraint
$P(x_i,y_i)\ \le\ P({y_i}_{x_i})\ \le\ 1-P(x_i)+P(x_i,y_i)$.
For each $n$, we compute the average improvement of our bounds over the $10{,}000$ instances
and plot the resulting differences in Figure~\ref{pic3}.
Across all dimensions, the improvements are never negative.
Moreover, the average improvement decreases as $n$ increases, which is expected because
probability mass is spread over more categories, making typical values of
$P(x_i,y_i)$ and $P({y_i}_{x_i})$ smaller.

\paragraph{Per-instance comparisons.}
To further illustrate the improvement at the level of individual instances, we consider several
representative settings, $n\in\{3,4,5\}, k\in\{3,4,5\}$ (chosen for convenience of presentation).
For each setting, we generate $10{,}000$ compatible sample distributions and compute bounds by
both methods.
For sample $i$, let $[a_i,b_i]$ denote the bounds from our theorem and $[c_i,d_i]$ denote the
bounds from Li and Pearl~\shortcite{li2024probabilities}.
We summarize the following quantities:
\begin{itemize}
    \item Average increase in the lower bound : $\frac{\sum{(a_i - c_i)}}{10000}$
    \item Average decrease in the upper bound : $\frac{\sum{(d_i - b_i)}}{10000}$
    \item Average gap from Li and Pearl: $\frac{\sum{(d_i - c_i)}}{10000}$
    \item Average gap from theorem 2 : $\frac{\sum{(b_i - a_i)}}{10000}$
    \item Count of sample distributions where Theorem~2 yields narrower bounds : $\sum f_i$ where, $f_i = 1$ if $(a_i > c_i)$ or $(b_i < d_i)$ and $f_i = 0$ otherwise.
\end{itemize}

Table~\ref{summary_dim} reports these summary statistics across different dimensions. Notably, the bounds are narrowed in approximately 10\%–20\% of the samples.

\begin{table}[!t]
\centering
\setlength{\tabcolsep}{3pt}
\renewcommand{\arraystretch}{1.05}
\begin{tabular}{cccccc}
\toprule
\makecell{$n,k$ } &
\makecell{Avg. inc.\\LB} &
\makecell{Avg. dec.\\UB} &
\makecell{Li--Pearl\\PNS gap} &
\makecell{Thm.~2\\PNS gap} &
\makecell{Bounds\\narrower} \\
\midrule
$3,3$ & 0.0002 & 0.0029 & 0.1691 & 0.1661 & 1737 \\
$4,4$ & 0.0000 & 0.0024 & 0.1006 & 0.0983 & 1652 \\
$5,5$ & 0.0000 & 0.0018 & 0.0669 & 0.0651 & 1628 \\
$5,4$ & 0.0000 & 0.0015 & 0.0789 & 0.0775 & 1186 \\
$4,3$ & 0.0000 & 0.0018 & 0.1257 & 0.1239 & 1088 \\
$5,3$ & 0.0000 & 0.0009 & 0.0952 & 0.0943 & 746 \\
\bottomrule
\end{tabular}

\caption{Comparison of bound tightening and PNS-gap reduction across dimensions over 10{,}000 sample distributions.}
\label{summary_dim}
\end{table}

\paragraph{Visualizing per-instance improvements.}
To complement the aggregate trends in Fig.~\ref{pic3} and the summary statistics in Table~\ref{summary_dim}, we further examine per-instance behavior to illustrate how bound tightening manifests at the level of individual samples across dimensions.

\begin{figure}[!b]
  \centering
  \includegraphics[width=0.85\linewidth]{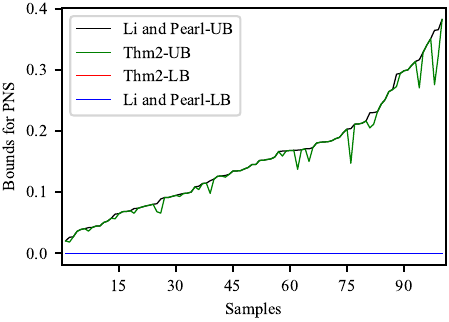}
  \caption{$\mathrm{PNS}$ bounds over 100 instances for $n=k=3$.}
  \label{pic_n00}
\end{figure}

\begin{figure*}[!t]
  \centering
  \begin{subfigure}[t]{0.3\textwidth}
    \centering
    \includegraphics[width=\linewidth]{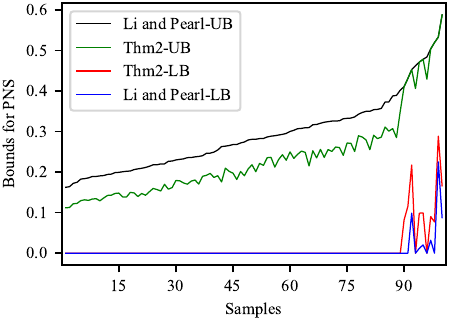}
    \caption{$\mathrm{PNS}$ bounds for $n=k=3$.}
    \label{pic_n3}
  \end{subfigure}\hfill
  \begin{subfigure}[t]{0.3\textwidth}
    \centering
    \includegraphics[width=\linewidth]{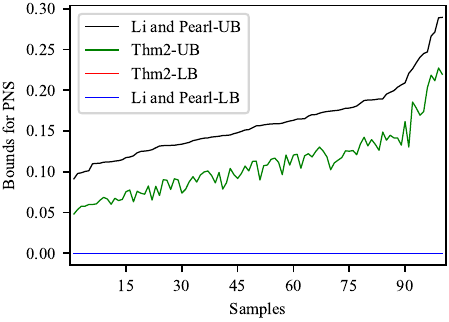}
    \caption{$\mathrm{PNS}$ bounds for $n=k=4$.}
    \label{pic_n4}
  \end{subfigure}\hfill
  \begin{subfigure}[t]{0.3\textwidth}
    \centering
    \includegraphics[width=\linewidth]{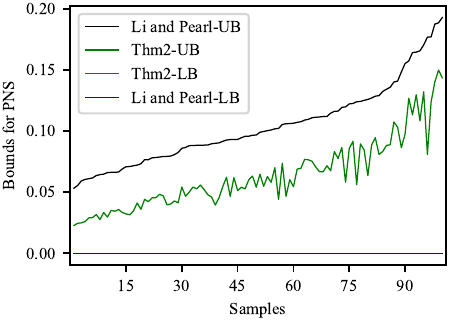}
    \caption{$\mathrm{PNS}$ bounds for $n=k=5$.}
    \label{pic_n5}
  \end{subfigure}

  \vspace{0.6em} 

  \begin{subfigure}[t]{0.3\textwidth}
    \centering
    \includegraphics[width=\linewidth]{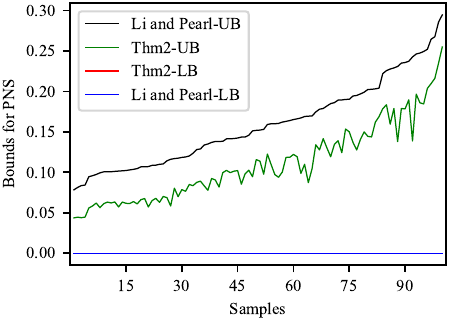}
    \caption{$\mathrm{PNS}$ bounds for $n=4,\,k=3$.}
    \label{pic_n4k3}
  \end{subfigure}\hfill
  \begin{subfigure}[t]{0.3\textwidth}
    \centering
    \includegraphics[width=\linewidth]{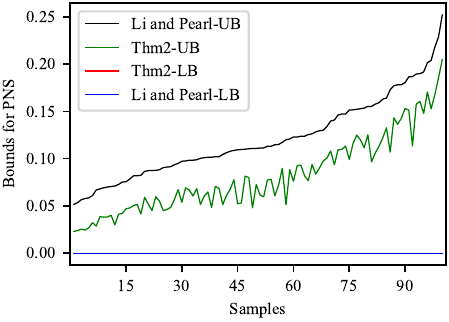}
    \caption{$\mathrm{PNS}$ bounds for $n=5,\,k=3$.}
    \label{pic_n5k3}
  \end{subfigure}\hfill
  \begin{subfigure}[t]{0.3\textwidth}
    \centering
    \includegraphics[width=\linewidth]{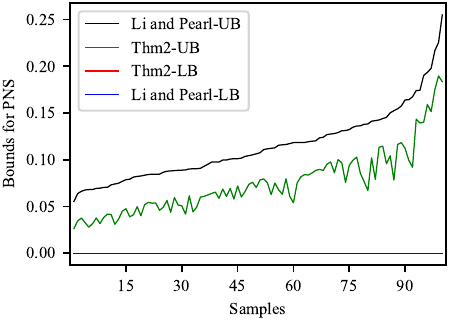}
    \caption{$\mathrm{PNS}$ bounds for $n=5,\,k=4$.}
    \label{pic_n5k4}
  \end{subfigure}

  \caption{Per-instance $\mathrm{PNS}$ bounds across different $(n,k)$ settings.}
  \label{fig:pns_345}
\end{figure*}

Figure~\ref{pic_n00} shows the lower and upper bounds for 100 representative instances with $n=k=3$, comparing the bounds from Theorem~2 and from Li and Pearl~\shortcite{li2024probabilities}. Extending this analysis, Fig.~\ref{fig:pns_345} reports results across multiple $(n,k)$ settings. For each configuration, we generate 10{,}000 compatible sample distributions and select the 100 instances with the largest discrepancies between the two methods, ordered by their PNS bounds to facilitate comparison.

As dimensionality increases, the lower bounds increasingly concentrate near zero. This behavior is expected, since probability mass is spread over a larger number of categories, reducing typical values of terms such as $P(x_i,y_i)$ and $P({y_i}_{x_i})$, and thereby limiting potential improvements in the lower bounds.

Importantly, this does not diminish the advantage of the proposed approach. Instead, with increasing $n$, the \emph{relative} separation between the bounds produced by Theorem~2 and those of Li and Pearl becomes more pronounced, driven mainly by systematic reductions in the upper bounds. This trend is particularly evident in Fig.~\ref{pic_n3}--\ref{pic_n5}, where $n=k$ and the relative gap widens as dimensionality grows.

Moreover, Fig.~\ref{pic_n4k3}--\ref{pic_n5k4} show that this separation persists when $n \neq k$, indicating that the closed-form bounds remain accurate beyond the diagonal ($n=k$) regime. Overall, these results demonstrate that the proposed closed-form expressions are not only computationally simpler, but also consistently tight across a wide range of dimensional configurations.

\section{Conclusion}
We studied identification of multi-valued probabilities of causation (PoCs) in general Structural Causal Models using standard experimental effects and observational joint distributions, without imposing structural restrictions. Our main contribution is a closed-form framework that reduces arbitrary discrete PoCs to a small family of representative conjunctions with analytically derived bounds. Two principles: equivalence classes and replaceability, enable systematic transfer of bounds across value permutations and mismatched treatment--outcome cardinalities, yielding a unified and compact toolkit for non-binary PoCs.

We established soundness for all dimensions and empirically verified tightness in low-dimensional settings by matching Balke’s linear programming solutions when feasible. Across simulations, our closed-form bounds consistently tighten the recursive bounds of Li and Pearl~\shortcite{li2024probabilities} while being substantially simpler to compute, making them well suited for repeated evaluation across many instances or queries. Based on these results, we conjecture that the proposed bounds are complete in general, although a formal proof for all dimensions remains open.

Our examples show that multi-valued PoCs capture cross-world response patterns beyond marginal comparisons. In the medical example, this supports risk-aware and heterogeneity-aware treatment assessment, while in the educational example it enables personalized decision making by quantifying risks under intensified interventions. More broadly, their value lies in revealing latent response types rather than average effects, thereby informing robust policies.

Several open directions remain. While tightness is verified in low dimensions, a general completeness proof is still open. Incorporating covariates and explicit causal graphs may further tighten bounds, and understanding when these additions sharpen identification is an important next step. Finally, although monotonicity yields point identification in some binary cases, its non-binary counterparts typically do not; identifying interpretable structural conditions that lead to sharper quantities remains challenging. Integrating Monotonic Incremental Treatment Effect (MITE)~\cite{mueller2025personalized} is a promising direction for deriving refined analytic bounds under partial-order constraints.

Overall, this work advances the theory of non-binary probabilities of causation by providing closed-form, sound, and practically computable bounds, enabling efficient and interpretable counterfactual attribution in multi-valued settings.

\bibliographystyle{named}
\bibliography{ijcai26}

\include{appendix}

\end{document}

%% file: appendix.tex
\onecolumn
\title{Supplementary Material}
\maketitle

\appendix
\section{Supplemental material}
\subsection{Probability of necessity and sufficiency($k$)}\label{nnkproof}
\begin{reptheorem}{nnk}[Probability of necessity and sufficiency($k$) (PNS($k$))]
Suppose variable $X$ has $n$ values $x_1,...,x_n$ and $Y$ has $n$ values $y_1,...,y_n$, $k \le n$, then the probability of necessity and sufficiency(k) $P({y_{1}}_{x_{1}},...,{y_{k}}_{x_{k}})$ is bounded as following:
\begin{eqnarray*}
\max \left \{
\begin{array}{cc}
0, \\
\displaystyle \sum_{j = 1}^{k}P({y_{j}}_{x_{j}}) - k + 1, \\
\displaystyle \sum_{\substack{1 \le j \le k \\ j \ne i}}\left[P({y_{j}}_{x_{j}})+P({x_{j}})-P({x_{j}, y_{j}})\right]+ \\
+ P({x_{i}, y_{i}}) - k + 1, & i \in \{1, ..., k\}
\end{array}
\right \}\nonumber
\le P({y_{1}}_{x_{1}},...,{y_{k}}_{x_{k}})
\end{eqnarray*}
\begin{eqnarray*}
\min \left \{
\begin{array}{cc}
\displaystyle \sum_{j = 1}^{k}P({x_{j}, y_{j}}) + \sum_{j = k+1}^{n}P({x_{j}}), \\
P({y_{j}}_{x_{j}}), & j \in \{1,...,k\},\\
\displaystyle \frac{1}{m}\left[\sum_{j = 0}^{m}P({y_{t_j}}_{x_{t_j}}) - P({x_{t_j}, y_{t_j}})\right], & m\in\{1,...,k-1\}, \\& t_j\in\{1,...,k\}
\end{array} 
\right \}\nonumber
\ge P({y_{1}}_{x_{1}},...,{y_{k}}_{x_{k}})
\end{eqnarray*}

\begin{proof}
By Fréchet Inequalities, we have,
\begin{eqnarray*}
 P(A_1,...,A_n) &\ge& \max\left\{0, P(A_1)+...+P(A_n)-n+1\right\},\\
 P(A_1,...,A_n) &\le& \min\left\{P(A_1),...,P(A_n)\right\}.
\end{eqnarray*}
Thus, we can easily derive the first two lower bounds and the second upper bound, 
\begin{eqnarray*}
P({y_{1}}_{x_{1}},...,{y_{k}}_{x_{k}})&\ge& \max\left\{0, \sum_{j = 1}^{k}P({y_{j}}_{x_{j}}) - k + 1\right\}\\
P({y_{1}}_{x_{1}},...,{y_{k}}_{x_{k}})&\le& \min_{1\le j \le k} \left\{P({y_{j}}_{x_{j}})\right\}.
\end{eqnarray*}
The equality of the first lower bound holds when $\exists j\in [1,k], \text{that } P({y_j}_{x_j})=0$.

The equality of the second lower bound holds when $\exists i \in [1,k]$, $P({y_j}_{x_j})=1$ for $\forall j \in [1,k], j\ne i$.

The equality of the second upper bound holds when $\exists j \in [1,k]$, $P({y_{i}}_{x_{i}}) = 1$ for $\forall i \in [1,k]$, $i\ne j$.

To proof the remaining lower bound $\sum_{\substack{1 \le j \le k \\ i \ne j}}\left[P({y_{j}}_{x_{j}})+P({x_{j}})-P({x_{j}, y_{j}})\right] + P({x_{i}, y_{i}}) - k + 1$, we start from
\begin{eqnarray*}
    P({y_{1}}_{x_{1}},...,{y_{k}}_{x_{k}}) 
    &=& \sum_{j=1}^{n} P({y_{1}}_{x_{1}},...,{y_{k}}_{x_{k}}, x_{j})\\
    &=& \sum_{j=1}^{k} P({y_{1}}_{x_{1}},...,{y_{k}}_{x_{k}}, x_{j}) + \sum_{j=k+1}^{n} P({y_{1}}_{x_{1}},...,{y_{k}}_{x_{k}}, x_{j})\\
    &\ge& \sum_{j=1}^{k} P({y_{1}}_{x_{1}},...,{y_{k}}_{x_{k}}, x_{j})\\
    &&\text{here, the equal sign holds when }\forall {x_j} = 0 \text{ for } j \in [k+1, n].
\end{eqnarray*}

For $\forall i, \text{s.t.}, i \in \{1, ..., k\}$, 
\begin{eqnarray*}
\sum_{j=1}^{k} P({y_{1}}_{x_{1}},...,{y_{k}}_{x_{k}}, x_{j}) &=& P({y_{1}}_{x_{1}},...,{y_{k}}_{x_{k}}, x_{i}) + \sum_{\substack{1 \le j \le k \\ i \ne j}}P({y_{1}}_{x_{1}},...,{y_{k}}_{x_{k}}, x_{j})\\
&\ge& P({y_{1}}_{x_{1}},...,{y_{k}}_{x_{k}}, x_{i})\\
&&\text{here, the equal sign holds when }\forall {x_j} = 0 \text{ for } j \in [1, k] \text{ and } j\ne i.
\end{eqnarray*}
thus, we have 
\begin{eqnarray*}
&& P({y_{1}}_{x_{1}},...,{y_{k}}_{x_{k}}) \\
&\ge& P({y_{1}}_{x_{1}},...,{y_{k}}_{x_{k}}, x_{i}) \\
&=& P({y_{1}}_{x_{1}},...,{y_{i-1}}_{x_{i-1}},{y_{i+1}}_{x_{i+1}},...,{y_{k}}_{x_{k}}, x_{i}, y_{i}) \\
&=& P({y_{1}}_{x_{1}},...,{y_{i-1}}_{x_{i-1}},{y_{i+1}}_{x_{i+1}},...,{y_{k}}_{x_{k}}, x_{i}, y_{i}) + \sum_{j=1}^{n}{P(x_{j})} - 1\\ 
&&+ \sum_{j=1}^{n} \sum_{q=1}^{n} P({y_{1}}_{x_{1}},...,{y_{i-1}}_{x_{i-1}},{y_{i+1}}_{x_{i+1}},...,{y_{k}}_{x_{k}}, x_{j}, y_{q}) \\
&&- \sum_{j=1}^{n} \sum_{q=1}^{n} P({y_{1}}_{x_{1}},...,{y_{i-1}}_{x_{i-1}},{y_{i+1}}_{x_{i+1}},...,{y_{k}}_{x_{k}}, x_{j}, y_{q}) \\
&=& P({y_{1}}_{x_{1}},...,{y_{i-1}}_{x_{i-1}},{y_{i+1}}_{x_{i+1}},...,{y_{k}}_{x_{k}}, x_{i}, y_{i}) \\
&&+ \sum_{\substack{1 \le j \le k \\ j \ne i}}{P(x_{j})} + P(x_{i}) + \sum_{j=k+1}^{n}{P(x_{j})} - 1 \\
&&+ P({y_{1}}_{x_{1}},...,{y_{i-1}}_{x_{i-1}},{y_{i+1}}_{x_{i+1}},...,{y_{k}}_{x_{k}}) \\
&&- \sum_{\substack{1 \le j \le k \\ j \ne i}}  P({y_{1}}_{x_{1}},...,{y_{i-1}}_{x_{i-1}},{y_{i+1}}_{x_{i+1}},...,{y_{k}}_{x_{k}}, x_{j}, y_{j}) \\
&&- \sum_{q=1}^{n} P({y_{1}}_{x_{1}},...,{y_{i-1}}_{x_{i-1}},{y_{i+1}}_{x_{i+1}},...,{y_{k}}_{x_{k}}, x_{i}, y_{q}) \\
&&- \sum_{j=k+1}^{n} \sum_{q=1}^{n} P({y_{1}}_{x_{1}},...,{y_{i-1}}_{x_{i-1}},{y_{i+1}}_{x_{i+1}},...,{y_{k}}_{x_{k}}, x_{j}, y_{q}) \\
&=& P({y_{1}}_{x_{1}},...,{y_{i-1}}_{x_{i-1}},{y_{i+1}}_{x_{i+1}},...,{y_{k}}_{x_{k}}) - 1 \\
&&+ \sum_{\substack{1 \le j \le k \\ j \ne i}}{P(x_{j})} - \sum_{\substack{1 \le j \le k \\ j \ne i}}  P({y_{1}}_{x_{1}},...,{y_{i-1}}_{x_{i-1}},{y_{i+1}}_{x_{i+1}},...,{y_{k}}_{x_{k}}, x_{j}, y_{j})\\
&&+ P(x_{i}) - \sum_{q=1}^{n} P({y_{1}}_{x_{1}},...,{y_{i-1}}_{x_{i-1}},{y_{i+1}}_{x_{i+1}},...,{y_{k}}_{x_{k}}, x_{i}, y_{q}) \\
&&+ P({y_{1}}_{x_{1}},...,{y_{i-1}}_{x_{i-1}},{y_{i+1}}_{x_{i+1}},...,{y_{k}}_{x_{k}}, x_{i}, y_{i}) \\
&&+ \sum_{j=k+1}^{n}{P(x_{j})} - \sum_{j=k+1}^{n} \sum_{q=1}^{n} P({y_{1}}_{x_{1}},...,{y_{i-1}}_{x_{i-1}},{y_{i+1}}_{x_{i+1}},...,{y_{k}}_{x_{k}}, x_{j}, y_{q})\\
&=& P({y_{1}}_{x_{1}},...,{y_{i-1}}_{x_{i-1}},{y_{i+1}}_{x_{i+1}},...,{y_{k}}_{x_{k}}) - 1 \\
&&+ \sum_{\substack{1 \le j \le k \\ j \ne i}}{P(x_{j})} - \sum_{\substack{1 \le j \le k \\ j \ne i}}  P({y_{1}}_{x_{1}},...,{y_{i-1}}_{x_{i-1}},{y_{i+1}}_{x_{i+1}},...,{y_{k}}_{x_{k}}, x_{j}, y_{j})\\
&&+ P(x_{i}) - \sum_{\substack{1 \le q \le n \\ q \ne i}} P({y_{1}}_{x_{1}},...,{y_{i-1}}_{x_{i-1}},{y_{i+1}}_{x_{i+1}},...,{y_{k}}_{x_{k}}, x_{i}, y_{q}) \\
&&+ \sum_{j=k+1}^{n}{P(x_{j})} - \sum_{j=k+1}^{n} P({y_{1}}_{x_{1}},...,{y_{i-1}}_{x_{i-1}},{y_{i+1}}_{x_{i+1}},...,{y_{k}}_{x_{k}}, x_{j})\\
&\ge& \sum_{\substack{1 \le j \le k \\ i \ne j}}{P({y_{j}}_{x_{j}})} - (k-2) - 1 + \sum_{\substack{1 \le j \le k \\ j \ne i}}{P(x_{j})} - \sum_{\substack{1 \le j \le k \\ j \ne i}}{P(x_{j}, y_{j})}\\
&&+ P(x_{i}) - \sum_{\substack{1 \le q \le n \\ q \ne i}}{P(x_{i}, y_{q})} + \sum_{j=k+1}^{n}{P(x_{j})} - \sum_{j=k+1}^{n}{P(x_{j})}\\
&&\text{here, the equal sign holds when }
\exists i \in [1,k], \text{ and } P({y_j}_{x_j})=1 \text{ for } \forall j \in [1,k], j\ne i.\\
&=& \sum_{\substack{1 \le j \le k \\ i \ne j}}\left[P({y_{j}}_{x_{j}})+P({x_{j}})-P({x_{j}, y_{j}})\right] + P({x_{i}, y_{i}}) - k + 1 \\
\end{eqnarray*}
The equality of the third lower bound holds when
\begin{itemize}
    \item $\forall {x_j} = 0$ for $j \in [1, n]$ and $j\ne i$
    \item $\text{and } \exists i \in [1,k], \text{ and } P({y_j}_{x_j})=1 \text{ for } \forall j \in [1,k], j\ne i.$
\end{itemize}

To proof the first upper bound, 
\begin{eqnarray*}
    P({y_{1}}_{x_{1}},...,{y_{k}}_{x_{k}}) &=& \sum_{j=1}^{n} P({y_{1}}_{x_{1}},...,{y_{k}}_{x_{k}}, x_{j})\\
    &=& \sum_{j=1}^{k} P({y_{1}}_{x_{1}},...,{y_{k}}_{x_{k}}, x_{j}) + \sum_{j=k+1}^{n} P({y_{1}}_{x_{1}},...,{y_{k}}_{x_{k}}, x_{j})\\
    &=& \sum_{j=1}^{k} P({y_{1}}_{x_{1}},...,{y_{j-1}}_{x_{j-1}}, {y_{j+1}}_{x_{j+1}}, ..., {y_{k}}_{x_{k}}, x_{j}, y_{j}) \\
    &&+ \sum_{j=k+1}^{n} P({y_{1}}_{x_{1}},...,{y_{k}}_{x_{k}}, x_{j})\\
    &\le& \sum_{j=1}^{k} P(x_{j}, y_{j}) + \sum_{j = k+1}^{n}P({x_{j}})
\end{eqnarray*}

The equality of the first upper bound holds when $P({y_j}_{x_j})=1 \text{ for } \forall j \in [1,k].$

To proof the remaining upper bound $ \frac{1}{m}\left[\sum_{j = 0}^{m}P({y_{t_j}}_{x_{t_j}}) - P({x_{t_j}, y_{t_j}})\right]$, for $m\in\{1,...,k-1\}$, $t_j\in\{1,...,k\}$, we can first write PNS($k$) into
\begin{eqnarray}
    P({y_{1}}_{x_{1}},...,{y_{k}}_{x_{k}}) &=& \sum_{j=1}^{n} P({y_{1}}_{x_{1}},...,{y_{k}}_{x_{k}}, x_{j}) \nonumber\\
    &=& \frac{1}{m} \times \left[(m) \sum_{j=1}^{n} P({y_{1}}_{x_{1}},...,{y_{k}}_{x_{k}}, x_{j})\right]  \label{pns(k)1/m*m}
\end{eqnarray}
For $m = 1$, 
\begin{eqnarray*}
    \frac{1}{m}\left[\sum_{j = 0}^{m}P({y_{t_j}}_{x_{t_j}}) - P({x_{t_j}, y_{t_j}})\right] 
    &=& \sum_{j = 0}^{1}P({y_{t_j}}_{x_{t_j}}) - P({x_{t_j}, y_{t_j}})\\
    &=& P({y_{t_0}}_{x_{t_0}}) + P({y_{t_1}}_{x_{t_1}}) - P({x_{t_0}, y_{t_0}}) - P({x_{t_1}, y_{t_1}})
\end{eqnarray*}
W.L.O.G., let $1 \le t_0 < t_1 \le k$ :
\begin{eqnarray}
    && P({y_{1}}_{x_{1}},...,{y_{k}}_{x_{k}}) \nonumber\\
    &=& \sum_{j=1}^{n} P({y_{1}}_{x_{1}},...,{y_{k}}_{x_{k}}, x_{j}) \nonumber\\
    &=& \sum_{j=1}^{k} P({y_{1}}_{x_{1}},...,{y_{k}}_{x_{k}}, x_{j}) + \sum_{j=k+1}^{n} P({y_{1}}_{x_{1}},...,{y_{k}}_{x_{k}}, x_{j}) \nonumber\\
    &\le& \sum_{j=1}^{t_0}{P({y_{t_0}}_{x_{t_0}}, x_j, {y_{j+t_1-t_0}}_{x_{j+t_1-t_0}})}
    + \sum_{j={t_0}+1}^{k-t_1+t_0} P({y_{t_0}}_{x_{t_0}}, x_{j}, {y_{j+t_1-t_0}}_{x_{j+t_1-t_0}}) \nonumber\\
    &&+ \sum_{j=1}^{t_1-t_0}{P({y_{t_0}}_{x_{t_0}}, x_{k-t_1+t_0+j}, {y_{j}}_{x_{j}})} + \sum_{j=k+1}^{n} P({y_{1}}_{x_{1}},...,{y_{k}}_{x_{k}}, x_{j}) \nonumber\\
    &&\text{here, the equal sign holds when }
    \exists 1\le t_0 < t_1 \le k, \nonumber\\
    &&\text{ and } P({y_i}_{x_i})=1 \text{ for } \forall i \in [1,k], i\ne t_0, i\ne j+t_1-t_0 \text{ for } j\in [1,k-t_1+t_0].\nonumber\\
    &=& \sum_{j=1}^{t_0}{P({y_{t_0}}_{x_{t_0}}, x_j, {y_{j+t_1-t_0}}_{x_{j+t_1-t_0}})} 
    + \sum_{j=t_0+1}^{k-t_1+t_0} P({y_{t_0}}_{x_{t_0}}, x_{j}, {y_{j+t_1-t_0}}_{x_{j+t_1-t_0}}) \nonumber\\
    &&+ \sum_{j=1}^{t_1-t_0}{P({y_{t_0}}_{x_{t_0}}, x_{k-t_1+t_0+j}, {y_{j}}_{x_{j}})} 
    + \sum_{j=k+1}^{n} P({y_{1}}_{x_{1}},...,{y_{k}}_{x_{k}}, x_{j}) \nonumber\\
    &&+ P({y_{t_0}}_{x_{t_0}}) + P({y_{t_1}}_{x_{t_1}}) - P({x_{t_0}, y_{t_0}}) - P({x_{t_1}, y_{t_1}}) \nonumber\\
    &&- \sum_{\substack{\{i_1,...,i_{t_0-1},i_{t_0+1},...,i_{k+2}\} \\ \in \{1,...,n\}^{k+1}}}{P({y_{i_1}}_{x_1},...,{y_{i_{t_0-1}}}_{x_{t_0-1}},{y_{t_0}}_{x_{t_0}},{y_{i_{t_0+1}}}_{x_{t_0+1}},...,{y_{i_{k}}}_{x_{k}},x_{i_{k+1}},y_{i_{k+2}})} \nonumber\\
    &&- \sum_{\substack{\{i_1,...,i_{t_1-1},i_{t_1+1},...,i_{k+2}\} \\ \in \{1,...,n\}^{k+1}}}{P({y_{i_1}}_{x_1},...,{y_{i_{t_1-1}}}_{x_{t_1-1}},{y_{t_1}}_{x_{t_1}},{y_{i_{t_1+1}}}_{x_{t_1+1}},...,{y_{i_{k}}}_{x_{k}},x_{i_{k+1}},y_{i_{k+2}})} \nonumber\\ 
    &&+ \sum_{\substack{\{i_1,...,i_{t_0-1},i_{t_0+1},...,i_{k}\} \\ \in \{1,...,n\}^{k-1}}}{P({y_{i_1}}_{x_1},...,{y_{i_{t_0-1}}}_{x_{t_0-1}},{y_{t_0}}_{x_{t_0}},{y_{i_{t_0+1}}}_{x_{t_0+1}},...,{y_{i_{k}}}_{x_{k}}, x_{t_0}, y_{t_0})} \nonumber\\
    &&+ \sum_{\substack{\{i_1,...,i_{t_1-1},i_{t_1+1},...,i_{k}\} \\ \in \{1,...,n\}^{k-1}}}{P({y_{i_1}}_{x_1},...,{y_{i_{t_1-1}}}_{x_{t_1-1}},{y_{t_1}}_{x_{t_1}},{y_{i_{t_1+1}}}_{x_{t_1+1}},...,{y_{i_{k}}}_{x_{k}}, x_{t_1}, y_{t_1})} \nonumber\\
    &=& P({y_{t_0}}_{x_{t_0}}) + P({y_{t_1}}_{x_{t_1}}) - P({x_{t_0}, y_{t_0}}) - P({x_{t_1}, y_{t_1}}) \nonumber\\
    &&+ \sum_{j=1}^{t_0-1}{P({y_{t_0}}_{x_{t_0}}, x_j, {y_{j+t_1-t_0}}_{x_{j+t_1-t_0}})}
    + {P({y_{t_0}}_{x_{t_0}}, x_{t_0}, {y_{t_1}}_{x_{t_1}})}  \nonumber\\
    &&+ \sum_{j={t_0}+1}^{k-t_1+t_0} P({y_{t_0}}_{x_{t_0}}, x_{j}, {y_{j+t_1-t_0}}_{x_{j+t_1-t_0}}) 
    + \sum_{j=1}^{t_1-t_0}{P({y_{t_0}}_{x_{t_0}}, x_{k-t_1+t_0+j}, {y_{j}}_{x_{j}})} \nonumber\\
    &&+ \sum_{j=k+1}^{n} P({y_{1}}_{x_{1}},...,{y_{k}}_{x_{k}}, x_{j}) \nonumber\\
    &&- \sum_{\substack{\{i_1,...,i_{t_0-1},i_{t_0+1},...,i_{k+2}\} \\ \in \{1,...,n\}^{k+1}}}{P({y_{i_1}}_{x_1},...,{y_{i_{t_0-1}}}_{x_{t_0-1}},{y_{t_0}}_{x_{t_0}},{y_{i_{t_0+1}}}_{x_{t_0+1}},...,{y_{i_{k}}}_{x_{k}},x_{i_{k+1}},y_{i_{k+2}})} \nonumber\\
    &&- \sum_{\substack{\{i_1,...,i_{t_1-1},i_{t_1+1},...,i_{k+2}\} \\ \in \{1,...,n\}^{k+1}}}{P({y_{i_1}}_{x_1},...,{y_{i_{t_1-1}}}_{x_{t_1-1}},{y_{t_1}}_{x_{t_1}},{y_{i_{t_1+1}}}_{x_{t_1+1}},...,{y_{i_{k}}}_{x_{k}},x_{i_{k+1}},y_{i_{k+2}})} \nonumber\\ 
    &&+ \sum_{\substack{\{i_1,...,i_{t_0-1},i_{t_0+1},...,i_{k}\} \\ \in \{1,...,n\}^{k-1}}}{P({y_{i_1}}_{x_1},...,{y_{i_{t_0-1}}}_{x_{t_0-1}},{y_{t_0}}_{x_{t_0}},{y_{i_{t_0+1}}}_{x_{t_0+1}},...,{y_{i_{k}}}_{x_{k}}, x_{t_0}, y_{t_0})} \nonumber\\
    &&+ \sum_{\substack{\{i_1,...,i_{t_1-1},i_{t_1+1},...,i_{k}\} \\ \in \{1,...,n\}^{k-1}}}{P({y_{i_1}}_{x_1},...,{y_{i_{t_1-1}}}_{x_{t_1-1}},{y_{t_1}}_{x_{t_1}},{y_{i_{t_1+1}}}_{x_{t_1+1}},...,{y_{i_{k}}}_{x_{k}}, x_{t_1}, y_{t_1})} \nonumber\\
    &=& P({y_{t_0}}_{x_{t_0}}) + P({y_{t_1}}_{x_{t_1}}) - P({x_{t_0}, y_{t_0}}) - P({x_{t_1}, y_{t_1}}) \nonumber\\
    &&- \sum_{\substack{\{i_1,...,i_{t_0-1},i_{t_0+1},...,i_{k+2}\} \\ \in \{1,...,n\}^{k+1}}}{P({y_{i_1}}_{x_1},...,{y_{i_{t_0-1}}}_{x_{t_0-1}},{y_{t_0}}_{x_{t_0}},{y_{i_{t_0+1}}}_{x_{t_0+1}},...,{y_{i_{k}}}_{x_{k}},x_{i_{k+1}},y_{i_{k+2}})} \nonumber\\
    &&+ \Bigg[\sum_{j=1}^{t_0-1}{P({y_{t_0}}_{x_{t_0}}, x_j, {y_{j+t_1-t_0}}_{x_{j+t_1-t_0}})} 
    + \sum_{j=t_0+1}^{k-t_1+t_0} P({y_{t_0}}_{x_{t_0}}, x_{j}, {y_{j+t_1-t_0}}_{x_{j+t_1-t_0}}) \nonumber\\
    &&+ \sum_{j=1}^{t_1-t_0}{P({y_{t_0}}_{x_{t_0}}, x_{k-t_1+t_0+j}, {y_{j}}_{x_{j}})} 
    + \sum_{j=k+1}^{n} P({y_{1}}_{x_{1}},...,{y_{k}}_{x_{k}}, x_{j}) \nonumber\\
    &&+ \sum_{\substack{\{i_1,...,i_{t_0-1},i_{t_0+1},...,i_{k}\} \\ \in \{1,...,n\}^{k-1}}}{P({y_{i_1}}_{x_1},...,{y_{i_{t_0-1}}}_{x_{t_0-1}},{y_{t_0}}_{x_{t_0}},{y_{i_{t_0+1}}}_{x_{t_0+1}},...,{y_{i_{k}}}_{x_{k}}, x_{t_0}, y_{t_0})} \Bigg] \nonumber\\
    &&- \sum_{\substack{\{i_1,...,i_{t_1-1},i_{t_1+1},...,i_{k+2}\} \\ \in \{1,...,n\}^{k+1}}}{P({y_{i_1}}_{x_1},...,{y_{i_{t_1-1}}}_{x_{t_1-1}},{y_{t_1}}_{x_{t_1}},{y_{i_{t_1+1}}}_{x_{t_1+1}},...,{y_{i_{k}}}_{x_{k}},x_{i_{k+1}},y_{i_{k+2}})} \nonumber\\
    &&+ \Bigg[{P({y_{t_0}}_{x_{t_0}}, x_{t_0}, {y_{t_1}}_{x_{t_1}})} \nonumber\\
    &&+ \sum_{\substack{\{i_1,...,i_{t_1-1},i_{t_1+1},...,i_{k}\} \\ \in \{1,...,n\}^{k-1}}}{P({y_{i_1}}_{x_1},...,{y_{i_{t_1-1}}}_{x_{t_1-1}},{y_{t_1}}_{x_{t_1}},{y_{i_{t_1+1}}}_{x_{t_1+1}},...,{y_{i_{k}}}_{x_{k}}, x_{t_1}, y_{t_1})}\Bigg] \label{m=1}\\
    &\le& P({y_{t_0}}_{x_{t_0}}) + P({y_{t_1}}_{x_{t_1}}) - P({x_{t_0}, y_{t_0}}) - P({x_{t_1}, y_{t_1}}) \label{t0t1}\\
    &&\text{here, the equal sign holds when }
    \exists 1\le t_0 < t_1 \le k, \text{ and } P({y_{t_0}}_{x_{t_0}})=0.\nonumber \\ 
    &&\text{(this is just a sufficient condition illustrating tightness, while the equality can also hold in other cases)}\nonumber
\end{eqnarray}

For $m = 2$, by applying equation (\ref{pns(k)1/m*m}), we can get:
\begin{eqnarray*}
    P({y_{1}}_{x_{1}},...,{y_{k}}_{x_{k}})
    &=& \frac{1}{2} \times \left[2 \sum_{j=1}^{n} P({y_{1}}_{x_{1}},...,{y_{k}}_{x_{k}}, x_{j})\right]\\
    &=& \frac{1}{2} \times \left[\sum_{j=1}^{n} P({y_{1}}_{x_{1}},...,{y_{k}}_{x_{k}}, x_{j}) + \sum_{j=1}^{n} P({y_{1}}_{x_{1}},...,{y_{k}}_{x_{k}}, x_{j})\right]\\
\end{eqnarray*}
Then W.L.O.G., let $1 \le t_0 < t_1 < t_2 \le k$. By applying equation (\ref{m=1}), we have
\begin{eqnarray*}
    &&P({y_{1}}_{x_{1}},...,{y_{k}}_{x_{k}}) \\
    &\le& \frac{1}{2} \times \Bigg\{\Bigg[P({y_{t_0}}_{x_{t_0}}) + P({y_{t_1}}_{x_{t_1}}) - P({x_{t_0}, y_{t_0}}) - P({x_{t_1}, y_{t_1}})\\
    &&- \sum_{\substack{\{i_1,...,i_{t_0-1},i_{t_0+1},...,i_{k+2}\} \\ \in \{1,...,n\}^{k+1}}}{P({y_{i_1}}_{x_1},...,{y_{i_{t_0-1}}}_{x_{t_0-1}},{y_{t_0}}_{x_{t_0}},{y_{i_{t_0+1}}}_{x_{t_0+1}},...,{y_{i_{k}}}_{x_{k}},x_{i_{k+1}},y_{i_{k+2}})} \\
    &&+ \Bigg[\sum_{j=1}^{t_0-1}{P({y_{t_0}}_{x_{t_0}}, x_j, {y_{j+t_1-t_0}}_{x_{j+t_1-t_0}})} 
    + \sum_{j=t_0+1}^{k-t_1+t_0} P({y_{t_0}}_{x_{t_0}}, x_{j}, {y_{j+t_1-t_0}}_{x_{j+t_1-t_0}}) \\
    &&+ \sum_{j=1}^{t_1-t_0}{P({y_{t_0}}_{x_{t_0}}, x_{k-t_1+t_0+j}, {y_{j}}_{x_{j}})} 
    + \sum_{j=k+1}^{n} P({y_{1}}_{x_{1}},...,{y_{k}}_{x_{k}}, x_{j}) \\
    &&+ \sum_{\substack{\{i_1,...,i_{t_0-1},i_{t_0+1},...,i_{k}\} \\ \in \{1,...,n\}^{k-1}}}{P({y_{i_1}}_{x_1},...,{y_{i_{t_0-1}}}_{x_{t_0-1}},{y_{t_0}}_{x_{t_0}},{y_{i_{t_0+1}}}_{x_{t_0+1}},...,{y_{i_{k}}}_{x_{k}}, x_{t_0}, y_{t_0})} \Bigg] \\
    &&- \sum_{\substack{\{i_1,...,i_{t_1-1},i_{t_1+1},...,i_{k+2}\} \\ \in \{1,...,n\}^{k+1}}}{P({y_{i_1}}_{x_1},...,{y_{i_{t_1-1}}}_{x_{t_1-1}},{y_{t_1}}_{x_{t_1}},{y_{i_{t_1+1}}}_{x_{t_1+1}},...,{y_{i_{k}}}_{x_{k}},x_{i_{k+1}},y_{i_{k+2}})} \\
    &&+ \bigg[{P({y_{t_0}}_{x_{t_0}}, x_{t_0}, {y_{t_1}}_{x_{t_1}})} \\
    &&+ \sum_{\substack{\{i_1,...,i_{t_1-1},i_{t_1+1},...,i_{k}\} \\ \in \{1,...,n\}^{k-1}}}{P({y_{i_1}}_{x_1},...,{y_{i_{t_1-1}}}_{x_{t_1-1}},{y_{t_1}}_{x_{t_1}},{y_{i_{t_1+1}}}_{x_{t_1+1}},...,{y_{i_{k}}}_{x_{k}}, x_{t_1}, y_{t_1})}\bigg]\Bigg] \\
    &&+ \Bigg[P({y_{t_2}}_{x_{t_2}}) - P({x_{t_2}, y_{t_2}})\\
    &&- \sum_{\substack{\{i_1,...,i_{t_2-1},i_{t_2+1},...,i_{k+2}\} \\ \in \{1,...,n\}^{k+1}}}{P({y_{i_1}}_{x_1},...,{y_{i_{t_2-1}}}_{x_{t_2-1}},{y_{t_2}}_{x_{t_2}},{y_{i_{t_2+1}}}_{x_{t_2+1}},...,{y_{i_{k}}}_{x_{k}},x_{i_{k+1}},y_{i_{k+2}})} \\ 
    &&+ \sum_{\substack{\{i_1,...,i_{t_2-1},i_{t_2+1},...,i_{k}\} \\ \in \{1,...,n\}^{k-1}}}{P({y_{i_1}}_{x_1},...,{y_{i_{t_2-1}}}_{x_{t_2-1}},{y_{t_2}}_{x_{t_2}},{y_{i_{t_2+1}}}_{x_{t_2+1}},...,{y_{i_{k}}}_{x_{k}}, x_{t_2}, y_{t_2})} \\
    &&+ \sum_{\substack{1 \le j \le k \\ j \ne t_2}}{P({y_{1}}_{x_{1}},...,{y_{k}}_{x_{k}}, x_{j})} + P({y_{1}}_{x_{1}},...,{y_{t_2-1}}_{x_{t_2+1}},{y_{t_2}}_{x_{t_2}},{y_{t_2+1}}_{x_{+1}},...,{y_{k}}_{x_{k}}, x_{t_2}, y_{t_2}) \\
    &&+ \sum_{j=k+1}^{n} P({y_{1}}_{x_{1}},...,{y_{k}}_{x_{k}}, x_{j})\Bigg]\Bigg\}\\
    &\le& \frac{1}{2} \times \Bigg\{\Bigg[P({y_{t_0}}_{x_{t_0}}) + P({y_{t_1}}_{x_{t_1}}) - P({x_{t_0}, y_{t_0}}) - P({x_{t_1}, y_{t_1}}) \\
    &&- \sum_{\substack{\{i_1,...,i_{t_0-1},i_{t_0+1},...,i_{k+2}\} \\ \in \{1,...,n\}^{k+1}}}{P({y_{i_1}}_{x_1},...,{y_{i_{t_0-1}}}_{x_{t_0-1}},{y_{t_0}}_{x_{t_0}},{y_{i_{t_0+1}}}_{x_{t_0+1}},...,{y_{i_{k}}}_{x_{k}},x_{i_{k+1}},y_{i_{k+2}})} \\
    &&+ \bigg[\sum_{j=1}^{t_0-1}{P({y_{t_0}}_{x_{t_0}}, x_j, {y_{j+t_1-t_0}}_{x_{j+t_1-t_0}})}
    + \sum_{j=t_0+1}^{k-t_1+t_0} P({y_{t_0}}_{x_{t_0}}, x_{j}, {y_{j+t_1-t_0}}_{x_{j+t_1-t_0}}) \\
    &&+ \sum_{j=1}^{t_1-t_0}{P({y_{t_0}}_{x_{t_0}}, x_{k-t_1+t_0+j}, {y_{j}}_{x_{j}})}
    + \sum_{j=k+1}^{n} P({y_{1}}_{x_{1}},...,{y_{k}}_{x_{k}}, x_{j}) \\
    &&+ \sum_{\substack{\{i_1,...,i_{t_0-1},i_{t_0+1},...,i_{k}\} \\ \in \{1,...,n\}^{k-1}}}{P({y_{i_1}}_{x_1},...,{y_{i_{t_0-1}}}_{x_{t_0-1}},{y_{t_0}}_{x_{t_0}},{y_{i_{t_0+1}}}_{x_{t_0+1}},...,{y_{i_{k}}}_{x_{k}}, x_{t_0}, y_{t_0})} \bigg]\\
    &&- \sum_{\substack{\{i_1,...,i_{t_1-1},i_{t_1+1},...,i_{k+2}\} \\ \in \{1,...,n\}^{k+1}}}{P({y_{i_1}}_{x_1},...,{y_{i_{t_1-1}}}_{x_{t_1-1}},{y_{t_1}}_{x_{t_1}},{y_{i_{t_1+1}}}_{x_{t_1+1}},...,{y_{i_{k}}}_{x_{k}},x_{i_{k+1}},y_{i_{k+2}})} \\
    &&+ \bigg[{P({y_{t_0}}_{x_{t_0}}, x_{t_0}, {y_{t_1}}_{x_{t_1}})} \\
    &&+ \sum_{\substack{\{i_1,...,i_{t_1-1},i_{t_1+1},...,i_{k}\} \\ \in \{1,...,n\}^{k-1}}}{P({y_{i_1}}_{x_1},...,{y_{i_{t_1-1}}}_{x_{t_1-1}},{y_{t_1}}_{x_{t_1}},{y_{i_{t_1+1}}}_{x_{t_1+1}},...,{y_{i_{k}}}_{x_{k}}, x_{t_1}, y_{t_1})}\bigg]\Bigg] \\
    &&+ \Bigg[P({y_{t_2}}_{x_{t_2}}) - P({x_{t_2}, y_{t_2}})\\
    &&- \sum_{\substack{\{i_1,...,i_{t_2-1},i_{t_2+1},...,i_{k+2}\} \\ \in \{1,...,n\}^{k+1}}}{P({y_{i_1}}_{x_1},...,{y_{i_{t_2-1}}}_{x_{t_2-1}},{y_{t_2}}_{x_{t_2}},{y_{i_{t_2+1}}}_{x_{t_2+1}},...,{y_{i_{k}}}_{x_{k}},x_{i_{k+1}},y_{i_{k+2}})} \\ 
    &&+ \sum_{\substack{\{i_1,...,i_{t_2-1},i_{t_2+1},...,i_{k}\} \\ \in \{1,...,n\}^{k-1}}}{P({y_{i_1}}_{x_1},...,{y_{i_{t_2-1}}}_{x_{t_2-1}},{y_{t_2}}_{x_{t_2}},{y_{i_{t_2+1}}}_{x_{t_2+1}},...,{y_{i_{k}}}_{x_{k}}, x_{t_2}, y_{t_2})} \\
    &&+ \sum_{\substack{1 \le j \le k \\ j \ne t_2}}{P({y_{1}}_{x_{1}},...,{y_{k}}_{x_{k}}, x_{j})} + {P({y_{t_1}}_{x_{t_1}}, x_{t_2}, {y_{t_2}}_{x_{t_2}})} + \sum_{j=k+1}^{n} P({y_{1}}_{x_{1}},...,{y_{k}}_{x_{k}}, x_{j})\Bigg]\Bigg\}\\
    &&\text{here, the equal sign holds when }\exists 1 \le t_1 < t_2 \le k, \forall P({y_j}_{x_j}) = 0 \text{ for } j \in [1, k], j\ne t_1.\\
    &=& \frac{1}{2} \times \Bigg\{\Bigg[P({y_{t_0}}_{x_{t_0}}) + P({y_{t_1}}_{x_{t_1}}) + P({y_{t_2}}_{x_{t_2}}) - P({x_{t_0}, y_{t_0}}) - P({x_{t_1}, y_{t_1}}) - P({x_{t_2}, y_{t_2}}) \\
    &&- \sum_{\substack{\{i_1,...,i_{t_0-1},i_{t_0+1},...,i_{k+2}\} \\ \in \{1,...,n\}^{k+1}}}{P({y_{i_1}}_{x_1},...,{y_{i_{t_0-1}}}_{x_{t_0-1}},{y_{t_0}}_{x_{t_0}},{y_{i_{t_0+1}}}_{x_{t_0+1}},...,{y_{i_{k}}}_{x_{k}},x_{i_{k+1}},y_{i_{k+2}})} \\
    &&+ \bigg[\sum_{j=1}^{t_0-1}{P({y_{t_0}}_{x_{t_0}}, x_j, {y_{j+t_1-t_0}}_{x_{j+t_1-t_0}})}
    + \sum_{j=t_0+1}^{k-t_1+t_0} P({y_{t_0}}_{x_{t_0}}, x_{j}, {y_{j+t_1-t_0}}_{x_{j+t_1-t_0}}) \\
    &&+ \sum_{j=1}^{t_1-t_0}{P({y_{t_0}}_{x_{t_0}}, x_{k-t_1+t_0+j}, {y_{j}}_{x_{j}})} + \sum_{j=k+1}^{n} P({y_{1}}_{x_{1}},...,{y_{k}}_{x_{k}}, x_{j}) \\
    &&+ \sum_{\substack{\{i_1,...,i_{t_0-1},i_{t_0+1},...,i_{k}\} \\ \in \{1,...,n\}^{k-1}}}{P({y_{i_1}}_{x_1},...,{y_{i_{t_0-1}}}_{x_{t_0-1}},{y_{t_0}}_{x_{t_0}},{y_{i_{t_0+1}}}_{x_{t_0+1}},...,{y_{i_{k}}}_{x_{k}}, x_{t_0}, y_{t_0})} \bigg]\\
    &&- \sum_{\substack{\{i_1,...,i_{t_1-1},i_{t_1+1},...,i_{k+2}\} \\ \in \{1,...,n\}^{k+1}}}{P({y_{i_1}}_{x_1},...,{y_{i_{t_1-1}}}_{x_{t_1-1}},{y_{t_1}}_{x_{t_1}},{y_{i_{t_1+1}}}_{x_{t_1+1}},...,{y_{i_{k}}}_{x_{k}},x_{i_{k+1}},y_{i_{k+2}})} \\
    &&+ \bigg[{P({y_{t_0}}_{x_{t_0}}, x_{t_0}, {y_{t_1}}_{x_{t_1}})} + {P({y_{t_1}}_{x_{t_1}}, x_{t_2}, {y_{t_2}}_{x_{t_2}})} \\
    &&+ \sum_{\substack{\{i_1,...,i_{t_1-1},i_{t_1+1},...,i_{k}\} \\ \in \{1,...,n\}^{k-1}}}{P({y_{i_1}}_{x_1},...,{y_{i_{t_1-1}}}_{x_{t_1-1}},{y_{t_1}}_{x_{t_1}},{y_{i_{t_1+1}}}_{x_{t_1+1}},...,{y_{i_{k}}}_{x_{k}}, x_{t_1}, y_{t_1})}\bigg]\Bigg] \\
    &&- \sum_{\substack{\{i_1,...,i_{t_2-1},i_{t_2+1},...,i_{k+2}\} \\ \in \{1,...,n\}^{k+1}}}{P({y_{i_1}}_{x_1},...,{y_{i_{t_2-1}}}_{x_{t_2-1}},{y_{t_2}}_{x_{t_2}},{y_{i_{t_2+1}}}_{x_{t_2+1}},...,{y_{i_{k}}}_{x_{k}},x_{i_{k+1}},y_{i_{k+2}})} \\ 
    &&+ \Bigg[\sum_{\substack{\{i_1,...,i_{t_2-1},i_{t_2+1},...,i_{k}\} \\ \in \{1,...,n\}^{k-1}}}{P({y_{i_1}}_{x_1},...,{y_{i_{t_2-1}}}_{x_{t_2-1}},{y_{t_2}}_{x_{t_2}},{y_{i_{t_2+1}}}_{x_{t_2+1}},...,{y_{i_{k}}}_{x_{k}}, x_{t_2}, y_{t_2})} \\
    &&+ \sum_{\substack{1 \le j \le k \\ j \ne t_2}}{P({y_{1}}_{x_{1}},...,{y_{k}}_{x_{k}}, x_{j})} + \sum_{j=k+1}^{n} P({y_{1}}_{x_{1}},...,{y_{k}}_{x_{k}}, x_{j})\Bigg]\Bigg\}\\
    &\le& \frac{1}{2} \times \left[P({y_{t_0}}_{x_{t_0}}) + P({y_{t_1}}_{x_{t_1}}) + P({y_{t_2}}_{x_{t_2}}) - P({x_{t_0}, y_{t_0}}) - P({x_{t_1}, y_{t_1}}) - P({x_{t_2}, y_{t_2}})\right]\\
    &&\text{here, the equal sign holds when }
    \exists 1\le t_0 < t_1 \le k, \text{ and } P({y_{t_0}}_{x_{t_0}})=P({y_{t_1}}_{x_{t_1}})=0.\\ 
    &&\text{(this is just a sufficient condition illustrating tightness, while the equality can also hold in other cases)}
\end{eqnarray*}

For $3 \le m \le k-1$, W.L.O.G., let $1 \le t_0 < ... < t_{m} \le k$. 
By applying equation (\ref{pns(k)1/m*m}), we can get
\begin{eqnarray*}
    P({y_{1}}_{x_{1}},...,{y_{k}}_{x_{k}}) 
    &=& \frac{1}{m} \times \left[\sum_{j=1}^{n} P({y_{1}}_{x_{1}},...,{y_{k}}_{x_{k}}, x_{j}) + (m-1) \sum_{j=1}^{n} P({y_{1}}_{x_{1}},...,{y_{k}}_{x_{k}}, x_{j})\right]\\
\end{eqnarray*}
By applying equation (\ref{t0t1}), we have
\begin{eqnarray*}
    P({y_{1}}_{x_{1}},...,{y_{k}}_{x_{k}})
    &\le& \frac{1}{m} \times \Bigg[P({y_{t_0}}_{x_{t_0}}) + P({y_{t_1}}_{x_{t_1}}) - P({x_{t_0}, y_{t_0}}) - P({x_{t_1}, y_{t_1}})\\ 
    &&+ (m-1) \sum_{j=1}^{n} P({y_{1}}_{x_{1}},...,{y_{k}}_{x_{k}}, x_{j})\Bigg]\\
\end{eqnarray*}
Similar to the proof when $m = 2$, we can derive results for $3\le m\le k-1$:
\begin{eqnarray*}
    P({y_{1}}_{x_{1}},...,{y_{k}}_{x_{k}})
    &\le& \frac{1}{m} \times \Bigg[P({y_{t_0}}_{x_{t_0}}) + P({y_{t_1}}_{x_{t_1}}) - P({x_{t_0}, y_{t_0}}) - P({x_{t_1}, y_{t_1}}) \\
    &&+ \sum_{j=2}^{m} \left[P({y_{t_j}}_{x_{t_j}}) - P({x_{t_j}, y_{t_j}})\right]\Bigg]\\
    &=& \frac{1}{m} \times \sum_{j=0}^{m} \left[P({y_{t_j}}_{x_{t_j}}) - P({x_{t_j}, y_{t_j}})\right]
\end{eqnarray*}
Overall, we have proved the third bound for $m\in [1,k-1]$.
\end{proof}
\end{reptheorem}

\subsection{Probability of substitute($k,p$)}\label{nnk+x_pproof}
\begin{reptheorem}{nnk+x_p}[Probability of substitute($k,p$) (PSub($k,p$))]
Suppose variable $X$ has $n$ values $x_1,...,x_n$ and $Y$ has $n$ values $y_1,...,y_n$, $k \le n$, then the probability $P({y_{1}}_{x_{1}},...,{y_{k}}_{x_{k}}, x_p)$, s.t., $p \ne j$ for $1\le j \le k$ is bounded as following:
\begin{eqnarray*}
\max \left \{
\begin{array}{cc}
0, \\
\displaystyle \sum_{j=1}^{k}\left[P({y_{j}}_{x_{j}})+P({x_{j}})-P({x_{j}, y_{j}})\right] + P({x_{p}}) - k
\end{array}
\right \}\nonumber
\le P({y_{1}}_{x_{1}},...,{y_{k}}_{x_{k}}, x_p)
\label{nnkxplb}
\end{eqnarray*}
\begin{eqnarray*}
\min \left \{
\begin{array}{cc}
P({x_{p}}),\\
P({y_{j}}_{x_{j}}) - P({x_{j}, y_{j}}), & j \in \{1,...,k\}\\
\end{array} 
\right \}\nonumber
\ge P({y_{1}}_{x_{1}},...,{y_{k}}_{x_{k}}, x_p)\\
\label{nnkxpub}
\end{eqnarray*}



\begin{proof}
By Fréchet Inequalities, we have,
\begin{eqnarray*}
 P(A_1,...,A_n) &\ge& 0,\\
 P(A_1,...,A_n) &\le& P(A_j), \text{ for } \forall 1 \le j \le n.
\end{eqnarray*}
Thus, we can obtain the first lower bound and the first upper bound, 
\begin{eqnarray*}
P({y_{1}}_{x_{1}},...,{y_{k}}_{x_{k}},{x_{p}})&\ge& 0\\
P({y_{1}}_{x_{1}},...,{y_{k}}_{x_{k}},{x_{p}})&\le& {P({x_{p}})}.
\end{eqnarray*}
The equality of the first lower bound holds when 
$\exists j\in [1,k], \text{that } P({y_j}_{x_j})=0$ or $x_p = 0$, $p\ne j$.

The equality of the first upper bound holds when $P({y_{j}}_{x_{j}}) = 1$ for $\forall j \in [1,k]$, $j\ne p$.

For the second lower bound
\begin{eqnarray*}
&&P({y_{1}}_{x_{1}},...,{y_{k}}_{x_{k}}, x_p)\\
&=& P({y_{1}}_{x_{1}},...,{y_{k}}_{x_{k}}, x_p) + \sum_{j=1}^{n}{P({y_{1}}_{x_{1}},...,{y_{k}}_{x_{k}}, x_j)} - \sum_{j=1}^{n}{P({y_{1}}_{x_{1}},...,{y_{k}}_{x_{k}}, x_j)}\\
&=& {P({y_{1}}_{x_{1}},...,{y_{k}}_{x_{k}})} + P({y_{1}}_{x_{1}},...,{y_{k}}_{x_{k}}, x_p) - \sum_{j=1}^{n}{P({y_{1}}_{x_{1}},...,{y_{k}}_{x_{k}}, x_j)}\\
&=& {P({y_{1}}_{x_{1}},...,{y_{k}}_{x_{k}})} + P({y_{1}}_{x_{1}},...,{y_{k}}_{x_{k}}, x_p) + \sum_{j=1}^{n}{P(x_j)} - 1 - \sum_{j=1}^{n}{P({y_{1}}_{x_{1}},...,{y_{k}}_{x_{k}}, x_j)}\\
&=& {P({y_{1}}_{x_{1}},...,{y_{k}}_{x_{k}})} - 1 + P({y_{1}}_{x_{1}},...,{y_{k}}_{x_{k}}, x_p) \\
&&+ \sum_{j=1}^{k}{P(x_j)} - \sum_{j=1}^{k}{P({y_{1}}_{x_{1}},...,{y_{k}}_{x_{k}}, x_j)} + \sum_{j=k+1}^{n}{P(x_j)} - \sum_{j=k+1}^{n}{P({y_{1}}_{x_{1}},...,{y_{k}}_{x_{k}}, x_j)}\\
&=& {P({y_{1}}_{x_{1}},...,{y_{k}}_{x_{k}})} - 1 \\
&&+ \sum_{j=1}^{k}{P(x_j)} - \sum_{j=1}^{k}{P({y_{1}}_{x_{1}},...,{y_{j-1}}_{x_{j-1}},{y_{j+1}}_{x_{j+1}},...,{y_{k}}_{x_{k}}, x_j, y_j)} \\
&&+ \sum_{j=k+1}^{n}{P(x_j)} - \sum_{j=k+1}^{n}{P({y_{1}}_{x_{1}},...,{y_{k}}_{x_{k}}, x_j)} + P({y_{1}}_{x_{1}},...,{y_{k}}_{x_{k}}, x_p)\\
&=& {P({y_{1}}_{x_{1}},...,{y_{k}}_{x_{k}})} - 1 \\
&&+ \sum_{j=1}^{k}{P(x_j)} - \sum_{j=1}^{k}{P({y_{1}}_{x_{1}},...,{y_{j-1}}_{x_{j-1}},{y_{j+1}}_{x_{j+1}},...,{y_{k}}_{x_{k}}, x_j, y_j)} \\
&&+ \sum_{j=k+1}^{n}{P(x_j)} - \sum_{\substack{k+1 \le j \le n, \\ j \ne p}}{P({y_{1}}_{x_{1}},...,{y_{k}}_{x_{k}}, x_j)} \\
&\ge& \sum_{j=1}^{k}{P({y_j}_{x_j})} - (k-1) - 1 + \sum_{j=1}^{k}{P(x_j)} - \sum_{j=1}^{k}{P(x_j, y_j)} + \sum_{j=k+1}^{n}{P(x_j)} - \sum_{\substack{k+1 \le j \le n, \\ j \ne p}}{P(x_j)}\\
&&\text{here, the equal sign holds when }
\exists j \in [1,n], \text{ and } P({y_i}_{x_i})=1 \text{ for } \forall i \in [1,k], i\ne j.\\
&=& \sum_{j=1}^{k}{P({y_j}_{x_j})} - k + \sum_{j=1}^{k}{P(x_j)} - \sum_{j=1}^{k}{P(x_j, y_j)} + P(x_p)\\
&=& \sum_{j=1}^{k}\left[P({y_{j}}_{x_{j}})+P({x_{j}})-P({x_{j}, y_{j}})\right] + P({x_{p}}) - k
\end{eqnarray*}
The equality of the second lower bound holds when $\exists j \in [1,n], P({y_i}_{x_i})=1 \text{ for } \forall i \in [1,k], i\ne j$.

For the remaining upper bounds, $\forall j \in [1,k]$:
\begin{eqnarray*}
&&P({y_{1}}_{x_{1}},...,{y_{k}}_{x_{k}}, x_p)\\
&=& P({y_{1}}_{x_{1}},...,{y_{k}}_{x_{k}}, x_p) + P({y_j}_{x_j}) - P({y_j}_{x_j})\\
&=& P({y_{1}}_{x_{1}},...,{y_{k}}_{x_{k}}, x_p) + P({y_j}_{x_j}) \\
&&- \sum_{\substack{\{i_1,...,i_{j-1},i_{j+1},...,i_{k+1}\} \\ \in \{1,...,n\}^{k}}}{P({y_{i_1}}_{x_1},...,{y_{i_{j-1}}}_{x_{j-1}},{y_{j}}_{x_{j}},{y_{i_{j+1}}}_{x_{j+1}},...,{y_{i_{k}}}_{x_{k}},x_{i_{k+1}})}
\end{eqnarray*}
Since $p \ne j$ for $1\le j \le k$, 
\begin{eqnarray*}
&&P({y_{1}}_{x_{1}},...,{y_{k}}_{x_{k}}, x_p)\\
&\le& P({y_j}_{x_j}) - \sum_{\substack{\{i_1,...,i_{j-1},i_{j+1},...,i_{k}\} \\ \in \{1,...,n\}^{k-1}}}{P({y_{i_1}}_{x_1},...,{y_{i_{j-1}}}_{x_{j-1}},{y_{j}}_{x_{j}},{y_{i_{j+1}}}_{x_{j+1}},...,{y_{i_{k}}}_{x_{k}},x_{j})}\\
&&\text{here, the equal sign holds when } P({x_j})=0 \text{ for } \forall j \in [k+1, n].\\
&=& P({y_j}_{x_j}) - P({y_{j}}_{x_{j}}, x_{j})\\
&=& P({y_{j}}_{x_{j}}) - P({x_{j}, y_{j}})
\end{eqnarray*}
The equality of the second upper bound holds when $P({x_j})=0 \text{ for } \forall j \in [k+1, n]$.
\end{proof}
\end{reptheorem}

\subsection{Probability of replacement($k,q$)}\label{nnk+y_qproof}
\begin{reptheorem}{nnk+y_q}[Probability of replacement($k,q$) (PRep($k,q$))]
Suppose variable $X$ has $n$ values $x_1,...,x_n$ and $Y$ has $n$ values $y_1,...,y_n$, $k \le n$, then the probability $P({y_{1}}_{x_{1}},...,{y_{k}}_{x_{k}}, y_q)$ is bounded as following:
\begin{eqnarray*}
\max \left \{
\begin{array}{cc}
0, \\
\displaystyle \sum_{j=1}^{k}\left[P({y_{j}}_{x_{j}})+P({x_{j}})-P({x_{j}, y_{j}})\right] +\\
+\displaystyle \sum_{\substack{k+1 \le j \le n \\ j \ne q}}{P({x_{j}, y_{q}})} + P({x_{q}, y_{q}}) - k,\\
\displaystyle \sum_{\substack{1 \le j \le k \\ j \ne q}}\left[P({y_{j}}_{x_{j}})+P({x_{j}})-P({x_{j}, y_{j}})\right] +\\
+ P({x_{q}, y_{q}}) - (k-1), & \text{if } q \in \{1,...,k\} 
\end{array}
\right \}\nonumber
\le P({y_{1}}_{x_{1}},...,{y_{k}}_{x_{k}}, y_q)
\label{nnkyqlb}
\end{eqnarray*}
\begin{eqnarray*}
\min \left \{
\begin{array}{cc}
P({y_{q}}_{x_{q}}), & \text{if } q \in \{1,...,k\}, \\
\displaystyle P({x_{q}, y_{q}}) + \sum_{\substack{k+1 \le j \le n \\ j \ne q}}P({x_{j}, y_{q}}),\\
P({y_{j}}_{x_{j}}) - P({x_{j}, y_{j}}), & j \in \{1,...,k\}, j \ne q\\
\end{array} 
\right \}\nonumber
\ge P({y_{1}}_{x_{1}},...,{y_{k}}_{x_{k}}, y_q)\\
\label{nnkyqub}
\end{eqnarray*}



\begin{proof}
By Fréchet Inequalities, we have,
\begin{eqnarray*}
P(A_1,...,A_n) &\ge& 0,\\
P(A_1,...,A_n) &\le& P(A_j), \text{ for } \forall 1 \le j \le n.
\end{eqnarray*}
Thus, we can obtain the first lower bound and the first upper bound, 
\begin{eqnarray*}
P({y_{1}}_{x_{1}},...,{y_{k}}_{x_{k}},{y_{q}})&\ge& 0\\
P({y_{1}}_{x_{1}},...,{y_{k}}_{x_{k}},{y_{q}})&\le& {P({y_{q}}_{x_{q}})}, \text{ for } \forall 1 \le q \le k.
\end{eqnarray*}
The equality of the first lower bound holds when
$\exists j\in [1,k], \text{that } P({y_j}_{x_j})=0$ or $y_q = 0$.

The equality of the first upper bound holds when $P({y_{j}}_{x_{j}}) = 1, y_q = 1$ for $\forall j \in [1,k], j\ne q$.

For the second lower bound
\begin{eqnarray*}
&&P({y_{1}}_{x_{1}},...,{y_{k}}_{x_{k}}, y_q)\\
&=& P({y_{1}}_{x_{1}},...,{y_{k}}_{x_{k}}, y_q) + P({y_{1}}_{x_{1}},...,{y_{k}}_{x_{k}}) - \sum_{j=1}^{n}\sum_{l=1}^{n}{P({y_{1}}_{x_{1}},...,{y_{k}}_{x_{k}}, x_j, y_l)}\\
&=& P({y_{1}}_{x_{1}},...,{y_{k}}_{x_{k}}, y_q) + P({y_{1}}_{x_{1}},...,{y_{k}}_{x_{k}}) \\
&&- \sum_{j=1}^{k}{P({y_{1}}_{x_{1}},...,{y_{k}}_{x_{k}}, x_j, y_j)} - \sum_{j=k+1}^{n}\sum_{l=1}^{n}{P({y_{1}}_{x_{1}},...,{y_{k}}_{x_{k}}, x_j, y_l)}\\
&=& P({y_{1}}_{x_{1}},...,{y_{k}}_{x_{k}}, y_q) + P({y_{1}}_{x_{1}},...,{y_{k}}_{x_{k}}) + \sum_{j=1}^{n}{P(x_j)} - 1\\
&&- \sum_{j=1}^{k}{P({y_{1}}_{x_{1}},...,{y_{k}}_{x_{k}}, x_j, y_j)} - \sum_{j=k+1}^{n}\sum_{l=1}^{n}{P({y_{1}}_{x_{1}},...,{y_{k}}_{x_{k}}, x_j, y_l)}\\
&=& \sum_{j=1}^{n}{P({y_{1}}_{x_{1}},...,{y_{k}}_{x_{k}}, x_j,y_q)} + P({y_{1}}_{x_{1}},...,{y_{k}}_{x_{k}}) - 1\\
&&+ \sum_{j=1}^{k}{P(x_j)} - \sum_{j=1}^{k}{P({y_{1}}_{x_{1}},...,{y_{k}}_{x_{k}}, x_j, y_j)} \\
&&+ \sum_{j=k+1}^{n}{P(x_j)} - \sum_{j=k+1}^{n}\sum_{l=1}^{n}{P({y_{1}}_{x_{1}},...,{y_{k}}_{x_{k}}, x_j, y_l)}
\end{eqnarray*}
If $q\in [1,k]$,
\begin{eqnarray*}
&&P({y_{1}}_{x_{1}},...,{y_{k}}_{x_{k}}, y_q)\\
&=& P({y_{1}}_{x_{1}},...,{y_{k}}_{x_{k}}) - 1\\
&&+ \sum_{j=1}^{k}{P(x_j)} - \sum_{j=1}^{k}{P({y_{1}}_{x_{1}},...,{y_{k}}_{x_{k}}, x_j, y_j)} + {P({y_{1}}_{x_{1}},...,{y_{k}}_{x_{k}}, x_q,y_q)} \\
&&+ \sum_{j=k+1}^{n}{P(x_j)} - \sum_{j=k+1}^{n}\sum_{l=1}^{n}{P({y_{1}}_{x_{1}},...,{y_{k}}_{x_{k}}, x_j, y_l)} + \sum_{j=k+1}^{n}{P({y_{1}}_{x_{1}},...,{y_{k}}_{x_{k}}, x_j,y_q)}\\
&=& P({y_{1}}_{x_{1}},...,{y_{k}}_{x_{k}}) - 1\\
&&+ \sum_{j=1}^{k}{P(x_j)} - \sum_{\substack{1\le j \le k,\\ j\ne q}}{P({y_{1}}_{x_{1}},...,{y_{k}}_{x_{k}}, x_j, y_j)}\\
&&+ \sum_{j=k+1}^{n}{P(x_j)} - \sum_{j=k+1}^{n}\sum_{\substack{1\le l \le n,\\l\ne q}}{P({y_{1}}_{x_{1}},...,{y_{k}}_{x_{k}}, x_j, y_l)} \\
&\ge& \sum_{j=1}^{k}{P({y_j}_{x_j})} - (k-1) - 1\\
&&+ \sum_{j=1}^{k}{P(x_j)} - \sum_{\substack{1\le j
\le k,\\ j\ne q}}{P(x_j, y_j)} + \sum_{j=k+1}^{n}{P(x_j)} - \sum_{j=k+1}^{n}\sum_{\substack{1\le l\le n,\\l\ne q}}{P(x_j, y_l)} \\
&&\text{here, the equal sign holds when }
\exists j \in [1,n], \text{ and } P({y_i}_{x_i})=1 \text{ for } \forall i \in [1,k], i\ne j.\\
&=& \sum_{j=1}^{k}{P({y_j}_{x_j})} - k + \sum_{j=1}^{k}{P(x_j)} - \sum_{j=1}^{k}{P(x_j, y_j)} + P(x_q, y_q) + \sum_{j=k+1}^{n}{P(x_j,y_q)}\\
&=& \sum_{j=1}^{k}\left[P({y_{j}}_{x_{j}})+P({x_{j}})-P({x_{j}, y_{j}})\right] + \sum_{\substack{k+1\le j \le n, \\ j \ne q}}{P({x_{j}, y_{q}})} + P({x_{q}, y_{q}}) - k
\end{eqnarray*}
If $q\in [k+1,n]$,
\begin{eqnarray*}
&&P({y_{1}}_{x_{1}},...,{y_{k}}_{x_{k}}, y_q)\\
&=& P({y_{1}}_{x_{1}},...,{y_{k}}_{x_{k}}) - 1\\
&&+ \sum_{j=1}^{k}{P(x_j)} - \sum_{j=1}^{k}{P({y_{1}}_{x_{1}},...,{y_{k}}_{x_{k}}, x_j, y_j)} \\
&&+ \sum_{j=k+1}^{n}{P(x_j)} - \sum_{j=k+1}^{n}\sum_{l=1}^{n}{P({y_{1}}_{x_{1}},...,{y_{k}}_{x_{k}}, x_j, y_l)} + \sum_{j=k+1}^{n}{P({y_{1}}_{x_{1}},...,{y_{k}}_{x_{k}}, x_j,y_q)}\\
&=& P({y_{1}}_{x_{1}},...,{y_{k}}_{x_{k}}) - 1 + \sum_{j=1}^{k}{P(x_j)} - \sum_{j=1}^{k}{P({y_{1}}_{x_{1}},...,{y_{k}}_{x_{k}}, x_j, y_j)}\\
&&+ \sum_{j=k+1}^{n}{P(x_j)} - \sum_{j=k+1}^{n}\sum_{\substack{1\le l\le n,\\l\ne q}}{P({y_{1}}_{x_{1}},...,{y_{k}}_{x_{k}}, x_j, y_l)}\\
&\ge& \sum_{j=1}^{k}{P({y_j}_{x_j})} - (k-1) - 1 + \sum_{j=1}^{k}{P(x_j)} - \sum_{j=1}^{k}{P(x_j, y_j)} \\
&&+ \sum_{j=k+1}^{n}{P(x_j)} - \sum_{j=k+1}^{n}\sum_{\substack{1\le l\le n,\\l\ne q}}{P(x_j, y_l)} \\
&&\text{here, the equal sign holds when }
\exists j \in [1,n], \text{ and } P({y_i}_{x_i})=1 \text{ for } \forall i \in [1,k], i\ne j.\\
&=& \sum_{j=1}^{k}{P({y_j}_{x_j})} - k + \sum_{j=1}^{k}{P(x_j)} - \sum_{j=1}^{k}{P(x_j, y_j)} + \sum_{j=k+1}^{n}{P({x_{j}, y_{q}})}\\
&=& \sum_{j=1}^{k}\left[P({y_{j}}_{x_{j}})+P({x_{j}})-P({x_{j}, y_{j}})\right] + \sum_{\substack{k+1\le j \le n, \\ j \ne q}}{P({x_{j}, y_{q}})} + P({x_{q}, y_{q}}) - k
\end{eqnarray*}
To summarize
\begin{eqnarray*}
&&P({y_{1}}_{x_{1}},...,{y_{k}}_{x_{k}}, y_q)\\
&\ge& \sum_{j=1}^{k}\left[P({y_{j}}_{x_{j}})+P({x_{j}})-P({x_{j}, y_{j}})\right] + \sum_{\substack{k+1 \le j \le n \\ j \ne q}}{P({x_{j}, y_{q}})} + P({x_{q}, y_{q}}) - k
\end{eqnarray*}
and the equality of the second lower bound holds when $\exists j \in [1,n], \text{ and } P({y_i}_{x_i})=1 \text{ for } \forall i \in [1,k], i\ne j$.

For the third lower bound, if $1\le q\le k$
\begin{eqnarray*}
&&P({y_{1}}_{x_{1}},...,{y_{k}}_{x_{k}}, y_q)\\
&=& \sum_{j=1}^{n}{P({y_{1}}_{x_{1}},...,{y_{k}}_{x_{k}}, x_j,y_q)}\\
&=& P({y_{1}}_{x_{1}},...,{y_{q-1}}_{x_{q-1}},{y_{q+1}}_{x_{q+1}},...,{y_{k}}_{x_{k}}, x_q,y_q) + \sum_{j=k+1}^{n}{P({y_{1}}_{x_{1}},...,{y_{k}}_{x_{k}}, x_j,y_q)}\\
&=& P({y_{1}}_{x_{1}},...,{y_{q-1}}_{x_{q-1}},{y_{q+1}}_{x_{q+1}},...,{y_{k}}_{x_{k}}, x_q,y_q) + \sum_{j=k+1}^{n}{P({y_{1}}_{x_{1}},...,{y_{k}}_{x_{k}}, x_j,y_q)} \\
&&+ P({y_{1}}_{x_{1}},...,{y_{q-1}}_{x_{q-1}},{y_{q+1}}_{x_{q+1}},...,{y_{k}}_{x_{k}}) \\
&&- \sum_{j=1}^{n}\sum_{l=1}^{n}{P({y_{1}}_{x_{1}},...,{y_{q-1}}_{x_{q-1}},{y_{q+1}}_{x_{q+1}},...,{y_{k}}_{x_{k}}, x_j, y_l)}\\
&&+ \sum_{j=1}^{n}{P(x_j)} - 1\\
&=& P({y_{1}}_{x_{1}},...,{y_{q-1}}_{x_{q-1}},{y_{q+1}}_{x_{q+1}},...,{y_{k}}_{x_{k}}) -1\\
&&+ \sum_{j=1}^{k}{P(x_j)} - \sum_{\substack{1\le j\le k,\\j\ne q}}{P({y_{1}}_{x_{1}},...,{y_{q-1}}_{x_{q-1}},{y_{q+1}}_{x_{q+1}},...,{y_{k}}_{x_{k}}, x_j, y_j)} \\
&&- \sum_{l=1}^{n}{P({y_{1}}_{x_{1}},...,{y_{q-1}}_{x_{q-1}},{y_{q+1}}_{x_{q+1}},...,{y_{k}}_{x_{k}}, x_q, y_l)} \\
&&+ P({y_{1}}_{x_{1}},...,{y_{q-1}}_{x_{q-1}},{y_{q+1}}_{x_{q+1}},...,{y_{k}}_{x_{k}}, x_q,y_q)\\
&&+ \sum_{j=k+1}^{n}{P(x_j)} - \sum_{j=k+1}^{n}\sum_{l=1}^{n}{P({y_{1}}_{x_{1}},...,{y_{q-1}}_{x_{q-1}},{y_{q+1}}_{x_{q+1}},...,{y_{k}}_{x_{k}}, x_j, y_l)} \\
&&+ \sum_{j=k+1}^{n}{P({y_{1}}_{x_{1}},...,{y_{k}}_{x_{k}}, x_j,y_q)} \\
&=& P({y_{1}}_{x_{1}},...,{y_{q-1}}_{x_{q-1}},{y_{q+1}}_{x_{q+1}},...,{y_{k}}_{x_{k}}) -1\\
&&+ \sum_{j=1}^{k}{P(x_j)} - \sum_{\substack{1\le j\le k,\\j\ne q}}{P({y_{1}}_{x_{1}},...,{y_{q-1}}_{x_{q-1}},{y_{q+1}}_{x_{q+1}},...,{y_{k}}_{x_{k}}, x_j, y_j)} \\
&&- \sum_{\substack{1\le l\le n,\\l\ne q}}{P({y_{1}}_{x_{1}},...,{y_{q-1}}_{x_{q-1}},{y_{q+1}}_{x_{q+1}},...,{y_{k}}_{x_{k}}, x_q, y_l)} \\
&&+ \sum_{j=k+1}^{n}{P(x_j)} - \sum_{j=k+1}^{n}\sum_{\substack{1\le l \le n,\\l\ne q}}{P({y_{1}}_{x_{1}},...,{y_{q-1}}_{x_{q-1}},{y_{q+1}}_{x_{q+1}},...,{y_{k}}_{x_{k}}, x_j, y_l)} \\
&\ge& \sum_{\substack{1\le j\le k,\\ j \ne q}}{P({y_{j}}_{x_{j}})} - (k-2) -1\\
&&+ \sum_{j=1}^{k}{P(x_j)} - \sum_{\substack{1\le j\le k,\\j\ne q}}{P(x_j, y_j)} - \sum_{\substack{1\le l \le n,\\l\ne q}}{P(x_q, y_l)} + \sum_{j=k+1}^{n}{P(x_j)} - \sum_{j=k+1}^{n}\sum_{\substack{1
\le l\le n,\\l\ne q}}{P(x_j, y_l)} \\
&&\text{here, the equal sign holds when }
\exists q \in [1,k], \text{ and } P({y_i}_{x_i})=1 \text{ for } \forall i \in [1,k], i\ne q.\\
&=& \sum_{\substack{1\le j \le k,\\ j \ne q}}{P({y_{j}}_{x_{j}})} - (k-1) \\
&&+ \sum_{\substack{1\le j\le k,\\j\ne q}}{P(x_j)} + P(x_q) - \sum_{\substack{1\le j\le k,\\j\ne q}}{P(x_j, y_j)} - \sum_{\substack{1\le l\le n,\\l\ne q}}{P(x_q, y_l)} + \sum_{j=k+1}^{n}{P(x_j, y_q)} \\
&=& \sum_{\substack{1\le j\le k, \\ j \ne q}}\left[P({y_{j}}_{x_{j}})+P({x_{j}})-P({x_{j}, y_{j}})\right] - (k-1)\\
&&+ P(x_q) - \sum_{\substack{1\le l\le n,\\l\ne q}}{P(x_q, y_l)} + \sum_{j=k+1}^{n}{P(x_j, y_q)} \\
&=& \sum_{\substack{1\le j\le k, \\ j \ne q}}\left[P({y_{j}}_{x_{j}})+P({x_{j}})-P({x_{j}, y_{j}})\right] - (k-1) + P(x_q, y_q) + \sum_{j=k+1}^{n}{P(x_j, y_q)} \\
&\ge& \sum_{\substack{1\le j\le k, \\ j \ne q}}\left[P({y_{j}}_{x_{j}})+P({x_{j}})-P({x_{j}, y_{j}})\right] + P({x_{q}, y_{q}}) - (k-1)\\
&&\text{here, the equal sign holds when }P({x_j})=0 \text{ for } \forall j \in [k+1,n].
\end{eqnarray*}
The equality of the third upper bound holds when $\exists q \in [1,k], \text{ and } P({y_i}_{x_i})=1, P({x_j})=0 \text{ for } \forall i \in [1,k], j\in [k+1, n], i\ne q$.

For the second upper bound
\begin{eqnarray*}
P({y_{1}}_{x_{1}},...,{y_{k}}_{x_{k}}, y_q) 
&=& \sum_{j=1}^{n}{P({y_{1}}_{x_{1}},...,{y_{k}}_{x_{k}}, x_j,y_q)}\\
&=& \sum_{j=1}^{k}{P({y_{1}}_{x_{1}},...,{y_{k}}_{x_{k}}, x_j,y_q)} + \sum_{j=k+1}^{n}{P({y_{1}}_{x_{1}},...,{y_{k}}_{x_{k}}, x_j,y_q)}
\end{eqnarray*}
If $q\in [1,k]$,
\begin{eqnarray*}
P({y_{1}}_{x_{1}},...,{y_{k}}_{x_{k}}, y_q) 
&=& P({y_{1}}_{x_{1}},...,{y_{k}}_{x_{k}}, x_q,y_q) + \sum_{j=k+1}^{n}{P({y_{1}}_{x_{1}},...,{y_{k}}_{x_{k}}, x_j,y_q)}\\
&\le& P(x_q,y_q) + \sum_{j=k+1}^{n}{P(x_j,y_q)}\\
&&\text{here, the equal sign holds when } P({y_i}_{x_i})=1 \text{ for } \forall i \in [1,k].\\
&=& P({x_{q}, y_{q}}) + \sum_{\substack{k+1\le j \le n, \\ j \ne q}}P({x_{j}, y_{q}})
\end{eqnarray*}
If $q\in [k+1,n]$,
\begin{eqnarray*}
P({y_{1}}_{x_{1}},...,{y_{k}}_{x_{k}}, y_q) 
&=& \sum_{j=k+1}^{n}{P({y_{1}}_{x_{1}},...,{y_{k}}_{x_{k}}, x_j,y_q)}\\
&\le& \sum_{j=k+1}^{n}{P(x_j,y_q)}\\
&&\text{here, the equal sign holds when } P({y_i}_{x_i})=1 \text{ for } \forall i \in [1,k].\\
&=& P({x_{q}, y_{q}}) + \sum_{\substack{k+1\le j \le n, \\ j \ne q}}P({x_{j}, y_{q}})
\end{eqnarray*}
To summarize
\begin{eqnarray*}
P({y_{1}}_{x_{1}},...,{y_{k}}_{x_{k}}, y_q) &\le& P({x_{q}, y_{q}}) + \sum_{\substack{k+1 \le j \le n \\ j \ne q}}P({x_{j}, y_{q}})
\end{eqnarray*}
and the equality of the second upper bound holds when $P({y_i}_{x_i})=1 \text{ for } \forall i \in [1,k]$.

For the remaining upper bounds, $\forall j \in [1,k]$:
\begin{eqnarray*}
&&P({y_{1}}_{x_{1}},...,{y_{k}}_{x_{k}}, y_q)\\
&=& \sum_{i=1}^{n}{P({y_{1}}_{x_{1}},...,{y_{k}}_{x_{k}}, x_i,y_q)} + P({y_j}_{x_j}) - P({y_j}_{x_j})\\
&=& \sum_{i=1}^{n}{P({y_{1}}_{x_{1}},...,{y_{k}}_{x_{k}}, x_i,y_q)} + P({y_j}_{x_j}) \\
&&- \sum_{\substack{\{i_1,...,i_{j-1},i_{j+1},...,i_{k+2}\} \\ \in \{1,...,n\}^{k+1}}}{P({y_{i_1}}_{x_1},...,{y_{i_{j-1}}}_{x_{j-1}},{y_{j}}_{x_{j}},{y_{i_{j+1}}}_{x_{j+1}},...,{y_{i_{k}}}_{x_{k}},x_{i_{k+1}},y_{i_{k+2}})}
\end{eqnarray*}
Since $q \ne j$ for $1\le j \le k$, 
\begin{eqnarray*}
&&P({y_{1}}_{x_{1}},...,{y_{k}}_{x_{k}}, y_q)\\
&\le& P({y_j}_{x_j}) - \sum_{\substack{\{i_1,...,i_{j-1},i_{j+1},...,i_{k}\} \\ \in \{1,...,n\}^{k-1}}}{P({y_{i_1}}_{x_1},...,{y_{i_{j-1}}}_{x_{j-1}},{y_{j}}_{x_{j}},{y_{i_{j+1}}}_{x_{j+1}},...,{y_{i_{k}}}_{x_{k}},x_{j},y_{j})}\\
&&\text{here, the equal sign holds when } P({x_j})=0,  P({y_j})=0 \text{ and } P({x_{t_1}, y_{t_2}})=0\\ &&\text{ for } \forall j \in [k+1, n], \forall t_1, t_2 \in [1, k] \text{ and } t_1\ne t_2.\\
&=& P({y_j}_{x_j}) - P({y_{j}}_{x_{j}}, x_{j}, y_{j})\\
&=& P({y_{j}}_{x_{j}}) - P({x_{j}, y_{j}})
\end{eqnarray*}
The equality of the third upper bound holds when $P({x_j})=0,  P({y_j})=0 \text{ and } P({x_{t_1}, y_{t_2}})=0 \text{ for } \forall j \in [k+1, n], t_1, t_2 \in [1, k] \text{ and } t_1\ne t_2$.
\end{proof}
\end{reptheorem}

\subsection{Probability of necessity($k,p,q$)}\label{nnk+x_p+y_qproof}
\begin{reptheorem}{nnk+x_p+y_q}[Probability of necessity($k,p,q$) (PN($k,p,q$))]
Suppose variable $X$ has $n$ values $x_1,...,x_n$ and $Y$ has $n$ values $y_1,...,y_n$, $k \le n$, then the probability $P({y_{1}}_{x_{1}},...,{y_{k}}_{x_{k}}, x_p, y_q)$, s.t., $p \ne j$ for $1\le j \le k$ is bounded as following:
\begin{eqnarray*}
\max \left \{
\begin{array}{cc}
0, \\
\displaystyle \sum_{j=1}^{k}\left[P({y_{j}}_{x_{j}})+P({x_{j}})-P({x_{j}, y_{j}})\right] + P({x_{p}, y_{q}}) - k\\
\end{array}
\right \}\nonumber
\le P({y_{1}}_{x_{1}},...,{y_{k}}_{x_{k}}, x_p, y_q)
\label{nnkxpyplb}
\end{eqnarray*}
\begin{eqnarray*}
\min \left \{
\begin{array}{cc}
P({x_{p}},{y_{q}}), \\
P({y_{j}}_{x_{j}}) - P({x_{j}, y_{j}}), & j \in \{1,...,k\} \\
\end{array} 
\right \}\nonumber
\ge P({y_{1}}_{x_{1}},...,{y_{k}}_{x_{k}}, x_p, y_q)
\label{nnkxpypub}
\end{eqnarray*}



\begin{proof}
By Fréchet Inequalities, we have,
\begin{eqnarray*}
 P(A_1,...,A_n) &\ge& 0,\\
 P(A_1,...,A_n) &\le& P(A_j), \text{ for } \forall 1 \le j \le n.
\end{eqnarray*}
Thus, we can obtain the first lower bound and the first upper bound, 
\begin{eqnarray*}
P({y_{1}}_{x_{1}},...,{y_{k}}_{x_{k}},{x_{p}},{y_{q}})&\ge& 0\\
P({y_{1}}_{x_{1}},...,{y_{k}}_{x_{k}},{x_{p}},{y_{q}})&\le& {P({x_{p}},{y_{q}})}.
\end{eqnarray*}
The equality of the first lower bound holds when 
$\exists j\in [1,k], \text{that } P({y_j}_{x_j})=0$ or $x_p = 0$ or $y_q = 0$, $p\ne j$.

The equality of the first upper bound holds when $P({y_{j}}_{x_{j}}) = 1$ for $\forall j \in [1,k]$.

For the second lower bound
\begin{eqnarray*}
&&P({y_{1}}_{x_{1}},...,{y_{k}}_{x_{k}}, x_p, y_q)\\
&=& P({y_{1}}_{x_{1}},...,{y_{k}}_{x_{k}}, x_p, y_q) + P({y_{1}}_{x_{1}},...,{y_{k}}_{x_{k}}) - P({y_{1}}_{x_{1}},...,{y_{k}}_{x_{k}})\\
&=& {P({y_{1}}_{x_{1}},...,{y_{k}}_{x_{k}})} + P({y_{1}}_{x_{1}},...,{y_{k}}_{x_{k}}, x_p, y_q) -\sum_{j=1}^{n}\sum_{l=1}^{n}{P({y_{1}}_{x_{1}},...,{y_{k}}_{x_{k}}, x_j, y_l)}\\
&=& {P({y_{1}}_{x_{1}},...,{y_{k}}_{x_{k}})} + P({y_{1}}_{x_{1}},...,{y_{k}}_{x_{k}}, x_p, y_q) + \sum_{j=1}^{n}{P(x_j)} - 1\\
&&- \sum_{j=1}^{k}{P({y_{1}}_{x_{1}},...,{y_{k}}_{x_{k}}, x_j, y_j)} - \sum_{j=k+1}^{n}\sum_{l=1}^{n}{P({y_{1}}_{x_{1}},...,{y_{k}}_{x_{k}}, x_j, y_l)} \\
&=& {P({y_{1}}_{x_{1}},...,{y_{k}}_{x_{k}})} - 1 \\
&&+ \sum_{j=1}^{k}{P(x_j)} - \sum_{j=1}^{k}{P({y_{1}}_{x_{1}},...,{y_{k}}_{x_{k}}, x_j, y_j)} \\ 
&&+ \sum_{j=k+1}^{n}{P(x_j)} - \sum_{j=k+1}^{n}\sum_{l=1}^{n}{P({y_{1}}_{x_{1}},...,{y_{k}}_{x_{k}}, x_j, y_l)} + P({y_{1}}_{x_{1}},...,{y_{k}}_{x_{k}}, x_p, y_q) \\
&=& {P({y_{1}}_{x_{1}},...,{y_{k}}_{x_{k}})} - 1\\
&&+ \sum_{j=1}^{k}{P(x_j)} - \sum_{j=1}^{k}{P({y_{1}}_{x_{1}},...,{y_{k}}_{x_{k}}, x_j, y_j)} \\ 
&&+ \sum_{j=k+1}^{n}{P(x_j)} - \sum_{\substack{k+1\le j \le n, \\ j \ne p}}\sum_{l=1}^{n}{P({y_{1}}_{x_{1}},...,{y_{k}}_{x_{k}}, x_j, y_l)} - \sum_{\substack{1\le l\le n, \\ l\ne p}}{P({y_{1}}_{x_{1}},...,{y_{k}}_{x_{k}}, x_p, y_l)}\\
&\ge& \sum_{j=1}^{k}{P({y_{j}}_{x_{j}})} - (k-1) - 1 \\
&&+ \sum_{j=1}^{k}{P(x_j)} - \sum_{j=1}^{k}{P(x_j, y_j)} + \sum_{j=k+1}^{n}{P(x_j)} - \sum_{\substack{k+1\le j \le n, \\ j \ne p}}{P(x_j)} - \sum_{\substack{1\le l \le n, \\ l\ne p}}{P(x_p, y_l)}\\
&&\text{here, the equal sign holds when }
\exists j \in [1,n], \text{ and } P({y_i}_{x_i})=1 \text{ for } \forall i \in [1,k], i\ne j.\\
&=& \sum_{j=1}^{k}{P({y_{j}}_{x_{j}})} - k + \sum_{j=1}^{k}{P(x_j)} - \sum_{j=1}^{k}{P(x_j, y_j)} + P(x_p) - \sum_{\substack{1\le l\le n, \\ l\ne p}}{P(x_p, y_l)}\\
&=& \sum_{j=1}^{k}\left[P({y_{j}}_{x_{j}})+P({x_{j}})-P({x_{j}, y_{j}})\right] + P({x_{p}, y_{q}}) - k
\end{eqnarray*}
The equality of the second lower bound holds when $\exists j \in [1,n], P({y_i}_{x_i})=1 \text{ for } \forall i \in [1,k], i\ne j$.

For the remaining upper bounds, $\forall j \in [1,k]$:
\begin{eqnarray*}
&&P({y_{1}}_{x_{1}},...,{y_{k}}_{x_{k}}, x_p, y_q)\\
&=& P({y_{1}}_{x_{1}},...,{y_{k}}_{x_{k}}, x_p, y_q) + P({y_j}_{x_j}) - P({y_j}_{x_j})\\
&=& P({y_{1}}_{x_{1}},...,{y_{k}}_{x_{k}}, x_p, y_q) + P({y_j}_{x_j}) \\
&&- \sum_{\substack{\{i_1,...,i_{j-1},i_{j+1},...,i_{k+2}\} \\ \in \{1,...,n\}^{k+1}}}{P({y_{i_1}}_{x_1},...,{y_{i_{j-1}}}_{x_{j-1}},{y_{j}}_{x_{j}},{y_{i_{j+1}}}_{x_{j+1}},...,{y_{i_{k}}}_{x_{k}},x_{i_{k+1}},y_{i_{k+2}})}
\end{eqnarray*}
Since $p \ne j$ for $1\le j \le k$, 
\begin{eqnarray*}
&&P({y_{1}}_{x_{1}},...,{y_{k}}_{x_{k}}, x_p, y_q)\\
&\le& P({y_j}_{x_j}) - \sum_{\substack{\{i_1,...,i_{j-1},i_{j+1},...,i_{k}\} \\ \in \{1,...,n\}^{k-1}}}{P({y_{i_1}}_{x_1},...,{y_{i_{j-1}}}_{x_{j-1}},{y_{j}}_{x_{j}},{y_{i_{j+1}}}_{x_{j+1}},...,{y_{i_{k}}}_{x_{k}},x_{j},y_{j})}\\
&&\text{here, the equal sign holds when } P({x_j})=0,  P({y_j})=0 \text{ and } P({x_{t_1}, y_{t_2}})=0\\ &&\text{ for } \forall j \in [k+1, n], \forall t_1, t_2 \in [1, k] \text{ and } t_1\ne t_2.\\
&=& P({y_j}_{x_j}) - P({y_{j}}_{x_{j}}, x_{j}, y_{j})\\
&=& P({y_{j}}_{x_{j}}) - P({x_{j}, y_{j}})
\end{eqnarray*}
The equality of the second upper bound holds when $P({x_j})=0,  P({y_j})=0 \text{ and } P({x_{t_1}, y_{t_2}})=0 \text{ for } \forall j \in [k+1, n], t_1, t_2 \in [1, k] \text{ and } t_1\ne t_2$.
\end{proof}
\end{reptheorem}

\subsection{Equivalence Class in Probability of Causation}
\begin{reptheorem}{nnk_mnk}[Equivalence classes in probabilities of causation]
Suppose variable $X$ has $n$ values $x_1,...,x_n$, $Y$ has $m$ values $y_1,...,y_m$:

\begin{itemize}
\item \textbf{Case 1:} For $k \le m<n$. Let $Y'$ have $n$ values $y'_1,...,y'_n$.  \\
Then the bounds of the probability, $P({y_{1}}_{x_{1}},...,{y_{k}}_{x_{k}})$, is exactly the same as the bounds of the probability, $P({y'_{1}}_{x_{1}},...,{y'_{k}}_{x_{k}})$, where
\begin{eqnarray*}
P({y'_{l}}_{x_j}) = 0, P(y'_l)=0, \text{ for }m+1 \le l \le n, 1 \le j \le n,
\end{eqnarray*}
\begin{eqnarray*}
\text{and, }P({y'_{l}}_{x_j}) = P({y_{l}}_{x_j}), P(x_j, y'_l)=P(x_j, y_l), \text{ for }1 \le l \le m, 1 \le j \le n.
\end{eqnarray*}
\item \textbf{Case 2:} For $k \le n<m$. Let $X'$ have $m$ values $x'_1,...,x'_m$.  \\
Then the bounds of the probability, $P({y_{1}}_{x_{1}},...,{y_{k}}_{x_{k}})$, is exactly the same as the bounds of the probability, $P({y_{1}}_{x'_{1}},...,{y_{k}}_{x'_{k}})$, where
\begin{eqnarray*}
P({y_{j}}_{x'_l}) = 0, P({y_{m}}_{x'_l}) = 1, P(x'_l) = 0, \text{ for }n+1 \le l \le m, 1 \le j \le m-1,
\end{eqnarray*}
\begin{eqnarray*}
\text{and, }
P({y_{j}}_{x'_l}) = P({y_{j}}_{x_l}), P(x'_l, y_j) = P(x_l, y_j) \text{ for }1 \le l \le n, 1 \le j \le m.
\end{eqnarray*}
\end{itemize}

\begin{proof}
\item \textbf{Case 1:}

Following \cite{tian2000probabilities}, the bounds of PNS are determined by a corresponding linear programming formulation. Therefore, to prove this theorem, we will show that the two probabilities share exactly the same linear programming formulation.

First, the bounds of $PNS$ of treatment $X$ with outcome $Y'$ are determined by the following Linear programming formulation:
\begin{equation*}
    max / min {\sum_{{j_{k+1},...,j_{n+1}}} {P_{1 ... k {j_{k+1}}... j_{n+1}}}}
\end{equation*}
and along with 
linear constraints:
\begin{eqnarray*}
    \sum_{j_1,...,j_{n+1}}{P_{j_1 ... j_{n+1}}} &=& 1\\
    P_{j_1 ... j_{n+1}} &\ge& 0
\end{eqnarray*}
$\forall s,t \in [1,n]$:\\
\begin{eqnarray*}
    \sum_{j_1,...j_{t-1},j_{t+1},...,j_{n+1}}{P_{j_1 ... j_{t-1} s j_{t+1}... j_{n+1}}} &=& P({y'_s}_{x_t})
\end{eqnarray*}
$\forall s,t \in [1,n]$:\\
\begin{eqnarray*}
    \sum_{j_1,...j_{t-1},j_{t+1},...,j_{n}}{P_{j_1 ... j_{t-1} s j_{t+1}... j_{n} t}} &=& P({x_t},{y'_s})
\end{eqnarray*}
Second, the bounds of $PNS$ of treatment $X$ with outcome $Y$ are determined by the following Linear programming formulation:
\begin{equation*}
    max / min {\sum_{j_{k+1},...,j_{n+1}}{P_{1 ... k {j_{k+1}}... j_{n+1}}}}
\end{equation*}
and along with 
linear constraints:
\begin{eqnarray*}
    \sum_{j_1,...,j_{n+1}}{P_{j_1 ... j_{n+1}}} &=& 1\\
    P_{j_1 ... j_{n+1}} &\ge& 0
\end{eqnarray*}
$\forall s \in [1,m], \forall t \in [1,n]$:\\
\begin{eqnarray*}
    \sum_{j_1,...j_{t-1},j_{t+1},...,j_{n+1}}{P_{j_1 ... j_{t-1} s j_{t+1}... j_{n+1}}} &=& P({y_s}_{x_t})
\end{eqnarray*}
$\forall s \in [1,m], \forall t \in [1,n]$:\\
\begin{eqnarray*}
    \sum_{j_1,...j_{t-1},j_{t+1},...,j_{n}}{P_{j_1 ... j_{t-1} s j_{t+1}... j_{n} t}} &=& P({x_t},{y_s})
\end{eqnarray*}
By setting 
\begin{eqnarray*}
    P({y'_{l}}_{x_j}) = 0, P(y'_l)=0
\end{eqnarray*}
for $m+1 \le l \le n$ and $1 \le j \le n$, and keeping  
\begin{equation*}
    P({y'_{l}}_{x_j}) = P({y_{l}}_{x_j}), P(x_j, y'_l)=P(x_j, y_l)
\end{equation*}
for $1 \le l \le m$ and $1 \le j \le n$, the two formulations are identical.

\item \textbf{Case 2:}

Again the bounds of $PNS$ of treatment $X'$ with outcome $Y$ are determined by the following Linear programming formulation, 
\begin{equation*}
    max / min {\sum_{j_{k+1},...,j_m,j_{n+1}}{P_{1 ... k {j_{k+1}}... j_m {n} ... {n} j_{n+1}}}}
\end{equation*}
and along with 
linear constraints:\\
\begin{eqnarray*}
    \sum_{j_1,...,j_m,j_{n+1}}{P_{j_1 ... j_m {n} ... {n} j_{n+1}}} &=& 1\\
    P_{j_1 ... j_m {n} ... {n} j_{n+1}} &\ge& 0
\end{eqnarray*}
$\forall s,t \in [1,m]$:\\
\begin{eqnarray*}
    \sum_{j_1,...,j_{t-1},j_{t+1}...j_m,j_{n+1}}{P_{j_1 ... j_{t-1} s j_{t+1} ... j_m {n} ... {n} j_{n+1}}} &=& P({y_s}_{x'_t})
\end{eqnarray*}
$\forall s,t \in [1,m]$:\\
\begin{eqnarray*}
    \sum_{j_1,...,j_{t-1},j_{t+1}...j_m}{P_{j_1 ... j_{t-1} s j_{t+1} ... j_m {n} ... {n} t}} &=& P({x'_t},{y_s})
\end{eqnarray*}

And the bounds of $PNS$ of treatment $X$ and outcome $Y$ are determined by the following Linear programming formulation:
\begin{equation*}
    max / min {\sum_{j_{k+1},...,j_{n+1}}{P_{1 ... k {j_{k+1}}...j_{n+1}}}}
\end{equation*}
and along with 
linear constraints:
\begin{eqnarray*}
    \sum_{j_1,...,j_{n+1}}{P_{j_1 ... j_{n+1}}} &=& 1\\
    P_{j_1 ... j_{n+1}} &\ge& 0
\end{eqnarray*}
$\forall s \in [1,m], \forall t \in [1,n]$:\\
\begin{eqnarray*}
    \sum_{j_1,...j_{t-1},j_{t+1},...,j_{n+1}}{P_{j_1 ... j_{t-1} s j_{t+1}... j_{n+1}}} &=& P({y_s}_{x_t})
\end{eqnarray*}
$\forall s \in [1,m], \forall t \in [1,n]$:\\
\begin{eqnarray*}
    \sum_{j_1,...j_{t-1},j_{t+1},...,j_{n}}{P_{j_1 ... j_{t-1} s j_{t+1}... j_{n} t}} &=& P({x_t},{y_s})
\end{eqnarray*}

By setting 
\begin{eqnarray*}
    P(x'_l) = 0, P({y_{j}}_{x'_l}) =0, P({y_{m}}_{x'_l}) = 1
\end{eqnarray*}
for $n+1 \le l \le m$ and $1 \le j \le m-1$
and keeping 
\begin{equation*}
    P({y_{j}}_{x'_l}) = P({y_{j}}_{x_l}), P(x'_l, y_j) = P(x_l, y_j)
\end{equation*}
for $1 \le l \le n, 1 \le j \le m$, the two formulations are again identical.

The proofs of equivalence for the other three theorems (Psub($K,p$), PRep($K,q$), and PN($K,p,q$)) can be derived using similar steps outlined above.
\end{proof}
\end{reptheorem}

\subsection{Replaceability in Probability of Causation}
\begin{reptheorem}{nnk_replacement}[Replaceability in probabilities of causation]
Suppose variable $X$ has $n$ values $x_1,...,x_n$ and $Y$ has $n$ values $y_1,...,y_n$, then the bounds of the probability, $P({y_{1}}_{x_{1}},...,{y_{i-1}}_{x_{i-1}},{y_{\hat{i}}}_{x_{i}},{y_{i+1}}_{x_{i+1}},...,{y_{k}}_{x_{k}})$,  can be obtained by replacing ${y_i}_{x_i}$ with ${y_{\hat{i}}}_{x_i}$ for any $i$, such that $1 \le i \le n$, in the bounds of the probability, $P({y_{1}}_{x_{1}},...,{y_{k}}_{x_{k}})$.

\begin{proof}
By replacing each term of ${y_i}_{x_i}$ with ${y_{\hat{i}}}_{x_i}$ in $P({y_{1}}_{x_{1}},...,{y_{k}}_{x_{k}})$, and following the same steps used in the proof of Theorem \ref{nnk} (see \ref{nnkproof}), we obtain exactly the same upper and lower bounds as the bounds of $P({y_{1}}_{x_{1}},...,{y_{i-1}}_{x_{i-1}},{y_{\hat{i}}}_{x_{i}},{y_{i+1}}_{x_{i+1}},...,{y_{k}}_{x_{k}})$.

The proofs of replaceability for the other three probabilities of causation (i.e., Psub($K,p$), PRep($K,q$), and PN($K,p,q$)) can be derived by replacing the term ${y_i}_{x_i}$ with ${y_{\hat{i}}}_{x_i}$ first and then following the same steps (as in \ref{nnk+x_pproof}, \ref{nnk+y_qproof}, \ref{nnk+x_p+y_qproof}) outlined above.
\end{proof}

\end{reptheorem}

\subsection{{Cross-Treatment Heterogeneity in Medicine}}\label{marketing}
The experimental data provide the following estimates:
\begin{eqnarray*}
&&P({y_1}_{x_1}) = 46 / 300, P({y_2}_{x_1}) = 23 / 300, 
P({y_3}_{x_1}) = 231 / 300,\\
&&P({y_1}_{x_2}) = 270 / 300, P({y_2}_{x_2}) = 8 / 300, P({y_3}_{x_2}) = 22 / 300,\\
&&P({y_1}_{x_3}) = 40 / 300, P({y_2}_{x_3}) = 223 / 300, 
P({y_3}_{x_3}) = 37 / 300.
\label{}
\end{eqnarray*}

The observational data provide the following estimates:
\begin{eqnarray*}
&&P(x_1,y_1) = 131 / 900,P(x_1,y_2) = 68 / 900, 
P(x_1,y_3) = 1 / 900,\\
&&P(x_2,y_1) = 45 / 900, P(x_2,y_2) = 22 / 900, 
P(x_2,y_3) = 51 / 900,\\
&&P(x_3,y_1) = 38 / 900, P(x_3,y_2) = 483 / 900, 
P(x_3,y_3) = 61 / 900.
\label{}
\end{eqnarray*}

By Theorem \ref{nnk} and Theorem \ref{nnk_replacement}, we derive the following bounds of the target probability of causation $P({y_{1}}_{x_{1}},{y_{2}}_{x_{2}},{y_{3}}_{x_{3}})$:


\begin{eqnarray*}
\max \left \{
\begin{array}{cc}
0, \\
\displaystyle P({y_{3}}_{x_{1}}) + P({y_{1}}_{x_{2}}) + P({y_{2}}_{x_{3}}) - 2, \\
P({y_{3}}_{x_{1}})+P({x_{1}})-P({x_{1}, y_{3}})\\
+P({y_{1}}_{x_{2}})+P({x_{2}})-P({x_{2}, y_{1}}) + P({x_{3}, y_{2}}) - 2,\\
P({y_{3}}_{x_{1}})+P({x_{1}})-P({x_{1}, y_{3}})\\
+P({y_{2}}_{x_{3}})+P({x_{3}})-P({x_{3}, y_{2}}) + P({x_{2}, y_{1}}) - 2,\\
P({y_{1}}_{x_{2}})+P({x_{2}})-P({x_{2}, y_{1}})\\
+P({y_{2}}_{x_{3}})+P({x_{3}})-P({x_{3}, y_{2}}) + P({x_{1}, y_{3}}) - 2
\end{array}
\right \}\nonumber
\le P({y_{3}}_{x_{1}},{y_{1}}_{x_{2}},{y_{2}}_{x_{3}})\\
\end{eqnarray*}

\begin{eqnarray*}
\min \left \{
\begin{array}{cc}
P({x_{1}, y_{3}}) + P({x_{2}, y_{1}}) + P({x_{3}, y_{2}}), \\
P({y_{3}}_{x_{1}}),\\
P({y_{1}}_{x_{2}}),\\
P({y_{2}}_{x_{3}}),\\
P({y_{3}}_{x_{1}}) + P({y_{1}}_{x_{2}}) - P({x_{1}, y_{3}}) - P({x_{2}, y_{1}}),\\
P({y_{3}}_{x_{1}}) + P({y_{2}}_{x_{3}}) - P({x_{1}, y_{3}}) - P({x_{3}, y_{2}}),\\
P({y_{1}}_{x_{2}}) + P({y_{2}}_{x_{3}}) - P({x_{2}, y_{1}}) - P({x_{3}, y_{2}}),\\
\frac{1}{2}\Big[P({y_{3}}_{x_{1}}) + P({y_{1}}_{x_{2}}) + P({y_{2}}_{x_{3}})\\ - P({x_{1}, y_{3}}) - P({x_{2}, y_{1}}) - P({x_{3}, y_{2}})\Big]
\end{array} 
\right \}\nonumber
\ge P({y_{3}}_{x_{1}},{y_{1}}_{x_{2}},{y_{2}}_{x_{3}})\\
\end{eqnarray*}

Thus,
\begin{eqnarray*}
\max \left \{
\begin{array}{cc}
0, \\
(231+270+223)/300 - 2, \\
(231+270)/300+(131+68+22+51+483)/900 - 2, \\
(231+223)/300+(131+68+38+61+45)/900 - 2,\\
(270+223)/300+(22+51+38+61+1)/900 - 2
\end{array}
\right \}\nonumber
\le P({y_{3}}_{x_{1}},{y_{1}}_{x_{2}},{y_{2}}_{x_{3}})\\
\end{eqnarray*}

\begin{eqnarray*}
\min \left \{
\begin{array}{cc}
(1+45+483)/900, \\
231/300,\\
270/300,\\
223/300,\\
(231+270)/300 - (1+45)/900,\\
(231+223)/300 - (1+483)/900,\\
(270+223)/300 - (45+483)/900,\\
\frac{1}{2}\Big[(231+270+223)/300 - (1+45+483)/900\Big]
\end{array} 
\right \}\nonumber
\ge P({y_{3}}_{x_{1}},{y_{1}}_{x_{2}},{y_{2}}_{x_{3}})\\
\end{eqnarray*}

Overall,
\begin{eqnarray*}
0.509\le P({y_3}_{x_1},{y_1}_{x_2},{y_2}_{x_3}) \le 0.588
\end{eqnarray*}

\subsection{{Personalized Decision-Making in Educational Interventions}}\label{supplement}

The experimental data provide the estimates:
\begin{eqnarray*}
&&P({y_1}_{x_1}) = 195 / 300,P({y_2}_{x_1}) = 51 / 300,P({y_3}_{x_1}) = 54 / 300,\\
&&P({y_1}_{x_2}) = 11 / 300,P({y_2}_{x_2}) = 266 / 300,P({y_3}_{x_2}) = 23 / 300,\\
&&P({y_1}_{x_3}) = 80 / 300,P({y_2}_{x_3}) = 198 / 300,P({y_3}_{x_3}) = 22 / 300,\\
&&P({y_1}_{x_4}) = 100 / 300,P({y_2}_{x_4}) = 147 / 300,P({y_3}_{x_4}) = 53 / 300
\label{}
\end{eqnarray*}

The observational data provide the estimates:
\begin{eqnarray*}
&&P(x_1,y_1) = 67 / 1200,P(x_1,y_2) = 129 / 1200,P(x_1,y_3) = 193 / 1200,\\
&&P(x_2,y_1) = 11 / 1200,P(x_2,y_2) = 17 / 1200,P(x_2,y_3) = 87 / 1200,\\
&&P(x_3,y_1) = 53 / 1200,P(x_3,y_2) = 53 / 1200,P(x_3,y_3) = 70 / 1200,\\
&&P(x_4,y_1) = 46 / 1200,P(x_4,y_2) = 436 / 1200,P(x_4,y_3) = 38 / 1200
\label{}
\end{eqnarray*}

We plug the experimental and observational estimates into Theorem \ref{nnk+x_p+y_q} and Theorem \ref{nnk_replacement} to obtain the following bounds:

\begin{eqnarray*}
\max \left \{
\begin{array}{cc}
0, \\
P({y_{1}}_{x_{1}})+P({x_{1}})-P({x_{1}, y_{1}}) + \\
+ P({y_{2}}_{x_{2}})+P({x_{2}})-P({x_{2}, y_{2}}) + \\
+ P({y_{2}}_{x_{3}})+P({x_{3}})-P({x_{3}, y_{2}}) + P({x_{4}, y_{2}}) - 3
\end{array}
\right \}\nonumber
\le P({y_1}_{x_1}, {y_2}_{x_2}, {y_2}_{x_3}, x_4, y_2)
\label{nnkxpyplb}
\end{eqnarray*}

\begin{eqnarray*}
\min \left \{
\begin{array}{cc}
P({x_{4}},{y_{2}}), \\
P({y_{1}}_{x_{1}}) - P({x_{1}, y_{1}}),\\
P({y_{2}}_{x_{2}}) - P({x_{2}, y_{2}}),\\
P({y_{2}}_{x_{3}}) - P({x_{3}, y_{2}})
\end{array} 
\right \}\nonumber
\ge P({y_1}_{x_1}, {y_2}_{x_2}, {y_2}_{x_3}, x_4, y_2)
\label{nnkxpypub}
\end{eqnarray*}

Thus,
\begin{eqnarray*}
\max \left \{
\begin{array}{cc}
0, \\
(195+266+198)/300+ \\+ (129+193+11+87+53+70+436)/1200 - 3
\end{array}
\right \}\nonumber
\le P({y_1}_{x_1}, {y_2}_{x_2}, {y_2}_{x_3}, x_4, y_2)
\label{nnkxpyplb}
\end{eqnarray*}

\begin{eqnarray*}
\min \left \{
\begin{array}{cc}
436/1200, \\
195/300 - 67/1200\\
266/300 - 17/1200\\
198/300 - 53/1200
\end{array} 
\right \}\nonumber
\ge P({y_1}_{x_1}, {y_2}_{x_2}, {y_2}_{x_3}, x_4, y_2)
\label{nnkxpypub}
\end{eqnarray*}

Overall,
\begin{eqnarray*}
0.0125 \le P({y_1}_{x_1}, {y_2}_{x_2}, {y_2}_{x_3}, x_4, y_2) \le 0.363
\end{eqnarray*}